\documentclass[runningheads]{llncs}
\usepackage{eccv}

\usepackage{eccvabbrv}
\usepackage{authblk}
\usepackage{graphicx}
\usepackage{booktabs}
\usepackage{wrapfig,lipsum}

\usepackage[accsupp]{axessibility}  % Improves PDF readability for those with disabilities.

\usepackage{amsmath,amsfonts,bm}

\def\eqref#1{equation~\ref{#1}}
\def\1{\bm{1}}

\DeclareMathAlphabet{\mathsfit}{\encodingdefault}{\sfdefault}{m}{sl}
\SetMathAlphabet{\mathsfit}{bold}{\encodingdefault}{\sfdefault}{bx}{n}

\def\gM{{\mathcal{M}}}

\def\gP{{\mathcal{P}}}

\def\gX{{\mathcal{X}}}

\newcommand{\OTsf}{\mathsf{OT}}
\newcommand{\UOTsf}{\mathsf{UOT}}
\DeclareMathOperator*{\argmin}{arg\,min}

\usepackage{hyperref}
\usepackage{orcidlink}
\usepackage{url}
\usepackage[ruled]{algorithm2e}
\usepackage{graphicx}
\usepackage{subcaption}
\usepackage{capt-of}
\usepackage{booktabs}
\usepackage{varwidth}
\newcommand{\minisection}[1]{\noindent{\textbf{#1}}}
\usepackage{blindtext}
\usepackage{booktabs}
\usepackage{tabularx}
\usepackage{footmisc}
\SetKw{kwInput}{Input:}
\SetKw{kwOutput}{Output:}

\begin{document}

% ---------------------------------------------------------------
% TODO REVIEW: Replace with your title

\title{A High-Quality Robust Diffusion Framework for Corrupted Dataset} 

% TODO REVIEW: If the paper title is too long for the running head, you can set
% an abbreviated paper title here. If not, comment out.
% \titlerunning{RDUOT}

% TODO FINAL: Replace with your author list. 
% Include the authors' OCRID for the camera-ready version, if at all possible.
\author{Quan Dao\inst{1,2}\textsuperscript{\textdagger \textdaggerdbl }\orcidlink{0009-0006-0996-0472} \and
Binh Ta\inst{1}\textsuperscript{\textdagger}\orcidlink{0000-0003-3553-5833} \and
Tung Pham\inst{1} \and
Anh Tran\inst{1}\orcidlink{0000-0002-3120-4036}
}

% TODO FINAL: Replace with an abbreviated list of authors.
\authorrunning{Quan Dao et al.}
% First names are abbreviated in the running head.
% If there are more than two authors, 'et al.' is used.

% TODO FINAL: Replace with your institution list.
\institute{
VinAI Research\\
\email{v.\{quandm7,binhth5,tungph4,anhtt152\}@vinai.io}\and
Rutgers University \\
\email{quan.dao@rutgers.edu}\\ 
}

\maketitle
\def\thefootnote{\textsuperscript{\textdaggerdbl}}\footnotetext{Work done during VinAI internship.}
\def\thefootnote{\textsuperscript{\textdagger}}\footnotetext{Equal contribution.}

\begin{abstract}

Developing image-generative models, which are robust to outliers in the training process, has recently drawn attention from the research community. Due to the ease of integrating unbalanced optimal transport (UOT) into adversarial framework, existing works focus mainly on developing robust frameworks for generative adversarial model (GAN). Meanwhile, diffusion models have recently dominated GAN in various tasks and datasets. However, according to our knowledge, none of them are robust to corrupted datasets. Motivated by DDGAN, our work introduces the first robust-to-outlier diffusion. We suggest replacing the UOT-based generative model for GAN in DDGAN to learn the backward diffusion process. Additionally, we demonstrate that the Lipschitz property of divergence in our framework contributes to more stable training convergence. Remarkably, our method not only exhibits robustness to corrupted datasets but also achieves superior performance on clean datasets.
\keywords{Diffusion Model \and Unbalanced Optimal Transport \and Robustness Generation \and OT-based Generative Model}
\end{abstract}

\section{Introduction} \label{sec_intro}
In recent years, generative models have seen remarkable advancements. These models have demonstrated the ability to generate pieces of writing, create stunning images, and even produce realistic videos in response to arbitrary queries. However, training datasets often originate from diverse sources, inevitably containing outliers resulting from various factors such as human error or machine inaccuracies. These outliers could significantly impede the performance of models; for instance, a generative model affected by outliers may produce undesired samples. In this study, we focus on a specific scenario where the training dataset for generative model is corrupted by outliers. 

The aforementioned scenario has been previously explored in the works \cite{Balaji_Robust, yang2018scalable}, primarily focusing on Generative Adversarial Networks (GANs). By leveraging unbalanced optimal transport (UOT), \cite{Balaji_Robust} proposed RobustGAN to enhance model robustness by using the third weight network to assign less attention to outliers and focus more on clean data. However, this approach not only requires additional training time and resources but also suffers from training instability due to the optimization of three networks, impairing the model's ability to create realistic images. Recently, \cite{rout2021generative} introduced OTM, a novel type of generative model known as the OT-based generative model, where the optimal transport map itself serves as a generative model. Building upon this work, \cite{choi2023generative} proposed UOTM framework which replaces the UOT formulation in the OT-based generative model. UOTM demonstrates strong performance on clean datasets, thereby bringing the OT-based generative model on par with other types of generative models such as diffusion and GANs in terms of quality. However, it is worth noting that UOTM only conducts robustness experiments on small-scale datasets with simplified settings, which may not accurately reflect real-world scenarios.

In addition to GANs, recent diffusion models \cite{ho2020denoising, song2019generative, song2020score, sohl2015deep} have experienced rapid growth due to their capability to outperform GANs in generating highly realistic images. These models offer adaptability in handling a wide range of conditional inputs, including semantic maps, text, and images, as highlighted in the works of \cite{rombach2022high, meng2021sdedit, wang2022semantic, ruiz2022dreambooth}. Despite these immense potentials, diffusion models face a significant weakness: slow sampling speed, as they require extremely large models with thousands of steps to slowly refine an image of white noise into a high-quality picture. This limitation impedes their widespread adoption, contrasting them with GANs. Hence, the combination of GANs and diffusion models, introduced in Denoising Diffusion GAN (DDGAN) \cite{xiao2022DDGAN}, has effectively addressed the challenge of modeling complex multimodal distributions, particularly when dealing with large step sizes, through the utilization of GANs. This innovative approach has led to a significant reduction in the number of denoising steps required, typically just a few (e.g., 2 or 4). On the other hand, robust generation is a critical issue frequently encountered in real-world scenarios. While this problem has been extensively studied in recent years, particularly in the context of GANs, it is evident that GANs still lag behind diffusion models in terms of image synthesis quality. Consequently, there is a growing consensus that diffusion models are poised to replace GANs as the leading approach in generative modeling. Given this shift in focus, it becomes imperative to address the question of how to train robust diffusion models that can effectively handle real-world datasets. To date, the development of robust diffusion models tailored for datasets containing a mixture of clean and outlier data points remains largely unexplored. Our work aims to fill this gap by proposing a robust diffusion framework capable of harnessing the high-quality synthesis capabilities of diffusion models while ensuring robustness throughout the generation process.

To address the challenge of producing a high-quality and fast sampling diffusion model in the presence of corrupted data, a straightforward solution might seem to be a combination of DDGAN and UOT, leveraging the strengths of both approaches. However, our work demonstrates that a simple combination of these techniques does not effectively solve the problem. Firstly, we demonstrate that DDGAN utilizes optimal transport (OT) to minimize the probability distance between fake and true distributions, whereas UOT learn to minimize the mapping between source and target distributions. Consequently, GAN and UOT have distinct objectives, making their direct combination challenging. Integrating UOTM into the GAN framework requires additional weight networks \cite{Balaji_Robust}, leading to poor convergence. In contrast, an OT-based generative framework \cite{rout2021generative} can seamlessly replace the UOT loss, as both share the same optimization objective. Motivated by this insight, we propose replacing the GAN process in DDGAN with an OT-based generative model to learn the backward diffusion process $q(x_{t-1}|x_t)$, facilitating the integration of the UOT loss. However, we discover that simply modeling $q(x_{t-1}|x_t)$ by UOT is ineffective because large $t$ makes it harder for UOT to distinguish between outliers and clean samples from $p(x_{t-1})$. To address this challenge, we propose learning the distribution $q(x_{0}|x_t)$ instead, as the UOT loss can more effectively filter out outliers from $q(x_0)$. Additionally, we highlight the effectiveness of Lipschitz $\Psi$ in stabilizing the training of the proposed framework. We summarize our contributions as follows:

\noindent$\bullet$ \textbf{Robust Diffusion UOT Framework}: We propose a novel approach to integrate UOT into the DDGAN framework by replacing the GAN process with an OT-based generative model. To address the challenge of distinguishing outliers from clean samples as diffusion steps increase, we propose to learn the distribution $p(x_{0}|x_t)$ instead of $q(x_{t-1}|x_t)$, leveraging the effectiveness of UOT in filtering outliers from the clean distribution $q(x_0)$.

\noindent $\bullet$ \textbf{Lipschitz $\Psi$ makes stable training}: We emphasize the importance of Lipschitz $\Psi$ in stabilizing the training process of our proposed framework, contributing to its overall effectiveness and stability.

\noindent $\bullet$ \textbf{Fast, High-fidelity, and Robust Image Generation}: Our proposed model exhibits superior performance compared to DDGAN and UOTM on clean datasets. Moreover, our framework demonstrates enhanced robustness, achieving a lower FID compared to other methods designed for robustness. 
%These results underscore the efficacy of our approach in generating high-quality images from clean datasets while also maintaining robustness against corrupted data.

\section{Related work}
In this section, we summarise the related works about unbalanced optimal transport (UOT) in generative models and diffusion models.

\textbf{UOT in generative models:} 
\cite{arjovsky2017wasserstein} proposed WGAN which showed the benefits of applying OT in GAN, which minimizes the Wasserstein distance between real and generated distribution. Indeed, OT theory has been the subject of extensive research over an extended period \cite{Villani2008OptimalTO, peyre2019computational, Cuturi-2013-Sinkhorn, altschuler2017near, janati2019spatio, pham2020-UOT}. This has led to techniques aimed at enhancing the efficiency of OT within GAN models \cite{sanjabi2018convergence, salimans2018improving}, all of which utilize Wasserstein distance.
Among the variants of OT, Unbalanced OT (UOT) has the potential to make a model more robust to training outliers \cite{gallouet2021regularity}. 
Recent works \cite{Balaji_Robust, yang2018scalable} proposed to integrate the UOT loss into GAN framework. However, these works need three distinct neural networks which leads to poor convergence and low-quality image synthesis. \cite{rout2021generative} proposed an OT-based generative model that optimal transport (OT) map itself can be used as a generative model. Recently, UOTM \cite{choi2023generative} extended OT-based generative model to UOT-based generative model by replacing OT formula with UOT formula. Though UOTM works well for clean datasets, its robustness experiments are only limited to low-resolution datasets. In this work, we show that our framework by extending the UOT-based generative model for diffusion framework achieves SoTA FID score at both clean and corrupted datasets.

\textbf{Diffusion models:} 
Diffusion models outperform state-of-the-art GANs in terms of high-fidelity synthesized images on various datasets \cite{dhariwal2021diffusion,saharia2022photorealistic}.
Furthermore, diffusion models also possess superior mode coverage \cite{song2021maximum, huang2021variational, kingma2021variational}, and offer adaptability in handling a wide range of conditional inputs including semantic maps, text, and images \cite{rombach2022high, meng2021sdedit, wang2022semantic}. This flexibility has led to their application in various areas, such as text-to-image generation, image-to-image translation, image inpainting, image restoration, and more \cite{ramesh2022hierarchical, saharia2022photorealistic, ruiz2022dreambooth, le_etal2023antidreambooth}. 
Nonetheless, their real-life application was shadowed by their slow sampling speed. DDPM \cite{ho2020denoising} requires a thousand sampling steps to obtain the high-fidelity image, resulting in long-time sampling. Although several techniques have been designed to reduce inference time \cite{song2020denoising, lu2022dpm, zhang2022fast}, primarily through reduction of sampling steps, they still need more than $10$ NFEs to generate images, roughly $10$ times slower than GANs. Recently, DDGAN \cite{xiao2022DDGAN} utilized GAN to tackle the challenge of modeling complex multimodal distributions caused by large step sizes. This model needs much fewer steps (e.g. 2 or 4) to generate an image.

\section{Background} \label{sec_background}

\subsection{Unbalanced Optimal Transport} \label{sec_background_uot}

In this section, we provide some background on optimal transport (OT), its unbalanced formulation (UOT), and its applications.\\
\textbf{Optimal Transport:} Let $\mu$ and $\nu$ be two probability measures in the set of probability measures $\gP(\gX)$ for space $\gX$, the OT distance between $\mu$ and $\nu$ is defined as:
\begin{align}
    \OTsf(\mu,\nu) = \min_{\pi \in \Pi(\mu,\nu)}\int c(x,y) d \pi(x,y),
\end{align}
where $c:\gX\times \gX\rightarrow [0,\infty)$ is a cost function,  $\Pi(\mu,\nu)$ is the set of joint probability measures on $\gX\times \gX$ which has $\mu$ and $\nu$ as marginal probability measures. The dual form of OT is:
\begin{align}
    \OTsf(\mu,\nu) &= \sup_{u(x) + v(y) \leq c(x,y)}\int_{\gX} u(x) d\mu(x)+ \int_{\gX}v(y) d\nu(y).
\end{align}
Denote $v^c(x) = \inf_{y\in \gX}\big\{ c(x,y) - v(y)\big\}$ to be the $c$-transform of $v(y)$, then the dual formulation of OT could be written in the following form:
\begin{align*}
    \OTsf(\mu,\nu) = \sup_{v}\int_{\gX}v^c(x) d\mu(x) + \int_{\gX}v(y) d\nu(y) .
\end{align*}
\textbf{Unbalanced Optimal Transport}:
A more generalized version of OT introduced by \cite{chizat2018unbalanced} is Unbalanced Optimal Transport (UOT) formulated as follows:
\begin{align} 
    \UOTsf(\mu,\nu) &= \min_{\pi \in \gM(\gX\times \gX)}\int \tau c(x,y) d\pi(x,y) + \mathsf{D}_{\Psi_1}(\pi_1\|\mu) + \mathsf{D}_{\Psi_2}(\pi_2\|\nu), \label{formulation:UOT}
\end{align}
where $\gM(\gX\times \gX)$ denotes the set of joint non-negative measures on $\gX\times \gX$; $\pi$ is an element of $\gM(\gX\times \gX)$, its marginal measures corresponding to $\mu$ and $\nu$ are $\pi_1$ and  $\pi_2$, respectively; the $\mathsf{D}_{\Psi_i}$ are often set as the Csisz\'{a}r-divergence, i.e., Kullback-Leibler divergence, $\chi^2$ divergence, $\tau$ is a hyper-parameter acting as the weight for the cost function. In contrast to OT, the UOT does not require hard constraints on the marginal distributions, thus allowing more flexibility to adapt to different situations. Similar to the OT, solving the UOT again could be done through its dual form \cite{chizat2018unbalanced, gallouet2021regularity, vacher2022stability}.
\begin{align}
\UOTsf(\mu, \nu)&=\sup _{u(x)+v(y) \leq \tau c(x, y)}\int_{\mathcal{X}}-\Psi_1^*(-u(x)) d \mu(x) + \int_{\mathcal{X}}-\Psi_2^*(-v(y)) d \nu(y), \label{eq_uot_dual}
\end{align}
where $u,v \in \mathcal{C}(\mathcal{X})$ in which  $\mathcal{C}$ denotes a set of continuous functions over its domain; $\Psi_1^*$ and $\Psi_2^*$ are the convex conjugate functions of $\Psi_1$ and $\Psi_2$, respectively.  
If both function $\Psi_1^*$ and $\Psi_2^*$ are non-decreasing and differentiable, we could next remove the condition $u(x) + v(y) \leq \tau c(x,y)$ by the $c$-transform for function $v$ to obtain the semi-dual UOT form \cite{vacher2022stability}, $v$ is 1-Lipschitz:
\begin{align}
\UOTsf(\mu,\nu) &= \sup_{||v||_{L} \leq 1} \int_{\gX} -\Psi_1^*\big(-v^c(x)\big) d\mu(x)  + \int_{\gX} -\Psi_2^*\big(-v(y) \big) d\nu(y). 
\label{eq_uot_semi}
\end{align}

Follow the definition of c-transform, UOTM \cite{choi2023generative} write $v^c(x) = \inf_{\hat{x} \in \gX} \tau c(x, \hat{x}) - v(\hat{x})$ where both optimal value of generated data $\hat{x}$ and potential function $v$ are unknown. Therefore, UOTM finds the function $v$ through learning a parameterized potential network $D_\phi$ and optimizing a parameterized generator $G_\theta: \mathcal{X} \rightarrow \mathcal{X}$ as mapping from input $x$ to the optimal value of $\hat{x}$. Therefore,  \cref{eq_uot_semi} can be written as follows:
\begin{align}
\UOTsf(\mu,\nu)
=&\sup_{D_\phi} \Big[\int_{\mathcal{X}} \Psi_1^*\Big(-\Big[\tau c\big(x, G_\theta(x)\big)-D_\phi\big(G_\theta(x)\big)\Big]\Big) d \mu(x) \nonumber \\
&+\int_{\mathcal{X}} \Psi_2^*\big(-D_\phi(y)\big) d \nu(y)\Big]\\
=&\inf _{D_\phi}\bigg[\int_{\mathcal{X}} \Psi_1^*\Big(-\inf _{G_\theta}\Big[\tau c\big(x, G_\theta(x)\big)-D_\phi\big(G_\theta(x)\big)\Big]\Big) d \mu(x) \nonumber \\
&+\int_{\mathcal{X}} \Psi_2^*\big(-D_{\phi}(y)\big) d \nu(y)\bigg]. \label{eq12}
\end{align}

\subsection{Diffusion Models} \label{sec_background_diff}
% Provide brief theories of diffusion
Diffusion models that rely on the diffusion process often take empirically thousand steps to diffuse the original data to become a neat approximation of Gaussian noise. Let's use $x_0$ to denote the true data, and $x_t$ denotes that datum after $t$ steps of rescaling data and adding Gaussian noise. The probability distributions of $x_t$ conditioned on $x_{t-1}$ and $x_0$ has the form
\begin{align}
q(x_t|x_{t-1}) &= \mathcal{N}(\sqrt{1-\beta_t}x_{t-1},\beta_t \mathbf{I}) 
 \label{xt_xt-1}\\
q(x_t|x_0) &= \mathcal{N}(x_t; \sqrt{\overline{\alpha}_t}x_0, (1 - \overline{\alpha}_t)\mathbf{I}) \label{xt}
\end{align}
where $\alpha_t = 1 - \beta_t$, $\overline{\alpha}_t = \prod_{s=1}^t \alpha_s$, and $\beta_t \in (0,1)$. Since the forward process introduces relatively minor noise each step, we can approximate reverse probability $p(x_{t-1}|x_t)$ using Gaussian probability $q(x_{t-1}|x_t,x_0)$, which could be learned through a parameterized function   $p_\theta(x_{t-1}|x_t)$. 
Following \cite{ho2020denoising},  $p_\theta(x_{t-1}|x_t)$ is commonly parameterized as:
\begin{equation}
    p_\theta(x_{t-1}\mid x_t) = \mathcal{N}(x_{t-1}; \mu_\theta(x_t, t), \sigma^2_t\mathbf{I}),
\end{equation}
where $\mu_\theta(x_t, t)$ and $\sigma^2_t$ represent the mean and variance of parameterized denoising model, respectively. The learning objective is to minimize the Kullback-Leibler (KL) divergence between true denoising distribution $q(x_{t-1}|x_t)$ and denoising distribution parameterized by $p_\theta(x_{t-1}|x_t)$. 

Unlike traditional methods, DDGAN \cite{xiao2022DDGAN} allows for larger denoising step sizes to speed up the sampling process by incorporating generative adversarial networks (GANs). DDGAN introduces a discriminator, denoted as $D_\phi$, and optimizes both the generator and discriminator in an adversarial training fashion. The objective of DDGAN can be expressed as follows:
\begin{align}
   \min_\phi \max_\theta &\sum_{t \geq 1} \mathbb{E}_{q\left(\mathbf{x}_t\right)}\bigg\{\mathbb{E}_{q\left(\mathbf{x}_{t-1} \mid \mathbf{x}_t\right)}\Big[-\log \big(D_\phi(\mathbf{x}_{t-1}, \mathbf{x}_t, t)\big)\Big] \nonumber \\
&+ \mathbb{E}_{p_\theta\left(\mathbf{x}_{t-1} \mid \mathbf{x}_t\right)}\Big[\log \big(D_\phi(\mathbf{x}_{t-1}, \mathbf{x}_t, t)\big)\Big]\bigg\}
    \label{eq:ddgan_obj_2}
\end{align}

In  \cref{eq:ddgan_obj_2}, conditional generator $p_\theta(x_{t-1} | x_t)$ generates fake samples. Due to large step sizes, the distribution $q(x_{t-1}|x_t)$ is no longer Gaussian. DDGAN models this complex multimodal distribution by using a generator $G_\theta(x_t, z, t)$, where $z$ is a $D$-dimensional latent variable drawn from a standard Gaussian distribution $\mathcal{N}(0, \mathbf{I})$. Specifically, DDGAN first generates an clean sample $x_0'$ through the generator $G_\theta(x_t, z, t)$ and obtains the perturbed sample $x_{t-1}'$ using $q(x_{t-1}|x_t,x'_0)$. Simultaneously, the discriminator evaluates both real pairs $D_\phi(x_{t-1}, x_t, t)$ and fake pairs $D_\phi(x_{t-1}', x_t, t)$ to guide the training process. 
%\cref{fig:ddgan_illustration} illustrate DDGAN training process.

\section{Method} \label{sec_method}

Recent works \cite{Balaji_Robust, choi2023generative} on robust generative models replace OT with UOT in adversarial framework. However, GANs are widely known for training instability and mode collapse \cite{kodali2017convergence}. By combining diffusion process and GAN models, Denoising Diffusion GAN (DDGAN) \cite{xiao2022DDGAN} successfully mitigates these limitations. While GAN uses OT distance to minimize the moving cost between real and fake distributions, UOT formulation minimizes the moving cost from source to target distributions. Therefore, it is hard to directly apply UOT into DDGAN framework. In the \cref{sec_robust_diffusion}, motivated by OT-based generative \cite{rout2021generative, choi2023generative}, we model backward diffusion process $p(x_{t-1}|x_t)$ by UOT-based generative model for robust-to-outlier image generation. However, naively modelling $p(x_{t-1}|x_t)$ leads to high FID since diffusion noising process reduces the difference between outlier and clean data. Instead, we model $p(x_{0}|x_t)$ by a UOT-based generative model to easily eliminate outliers. \cref{sec:analysis} presents the importance of Lipschizt property of $\Psi$ and how to design the potential network $D_\phi$, generator network $G_\theta$.

\subsection{Robust-to-Outlier Diffusion Framework}
\label{sec_robust_diffusion} 

DDGAN matches the conditional GAN generator $p_\theta(x_{t-1}|x_t)$ and $q(x_{t-1}|x_t)$ using an adversarial framework that minimizes OT loss per denoising step:
\begin{equation}
    \min _\theta \sum_{t \geq 1} \mathbb{E}_{q\left(x_t\right)}\OTsf\left(q\left(x_{t-1} \mid x_t\right) \| p_\theta\left(x_{t-1} \mid x_t\right)\right) \label{eq:ddgan_obj}
\end{equation}
where $q(x_{t-1}|x_t)$ is ground-truth conditional distribution with $x_{t-1}$ sampling from \cref{xt} and $x_{t}$ sampling from \cref{xt_xt-1}. The fake conditional pair $(\hat{x}_{t-1}, x_{t}) \sim p_\theta(x_{t-1}|x_t)$ is obtained using ground truth $x_t$ and $\hat{x}_{t-1} \sim q(x_{t-1}|x_t,\hat{x}_0)$ with $\hat{x}_0 = G_\theta(x_t, z, t)$ ($z \sim \mathcal{N}(0, \mathbb{I})$). Noted that: In DDGAN, \textbf{OT cost serves as the loss} to minimize that distance between true distribution $q(x_{t-1}|x_t)$ and fake distribution $p_\theta(x_{t-1}|x_t)$. For robustness problem, we cannot directly apply UOT formulation into GAN-based architecture since UOT does not measure the distance between true and fake distributions. To apply UOT in GAN, RobustGAN \cite{Balaji_Robust} needs additional network $W$ to weight the outliers, which leads to training instability due to optimization of three networks.

Motivated from \cite{choi2023generative, rout2021generative}, instead of minimizing OT cost between $q(x_{t-1}|x_t)$ and $p_\theta(x_{t-1}|x_t)$, our framework uses \textbf{optimal transport map as a generative model itself}, which is an OT-based generative model \cite{choi2023generative, rout2021generative}. To enable robustness property, we aim to learn a UOT mapping from marginal distribution $q(x_t)$ to backward diffusion process $q(x_{t-1}|x_t)$.

\begin{equation}
    \sum_{t \geq 1} \UOTsf\left(q(x_t) , q\left(x_{t - 1} |x_t\right)\right) \label{eq:DDUOT}
\end{equation}

However, due to diffusion process, the robustness property of generative model trained by \cref{eq:DDUOT} is not guaranteed. In \cref{formulation:UOT}, if $\tau$ is too small, UOT formulation becomes an OT formulation which penalizes the marginal constraints and ignores the outlier filtering. In contrast, when $\tau$ is too large, UOT formulation focus more to outlier filtering and ignores the marginal constraints. In case, the outlier and clean distributions are close to each other, $\tau$ should be increased for robustness guarantee. By \cref{prop:diffuse} (proof in Appendix 8), the outlier and clean noisy samples at time $t$ become close to each other as $t$ increases and \textbf{the $\tau$ should also increase as $t$ increases}. It is hard to cast out the outlier among $x_{t}$ since \textbf{choosing different $\tau$ for each step $t$ costs a huge amount of time and resource}. Furthermore, when the outlier and clean noisy samples for large $t$ are too similar, \textbf{large $\tau$ could accidentally remove the low-density modality of clean distribution and cannot eliminate the outlier samples}.

\begin{proposition}
\label{prop:diffuse}
Denote $P^c$ and $P^o$ be clean and outlier probability measures. Let $P_t$ be the probability measure that $x_t \sim P_t$ is obtained from $x_0 \sim P$ by a forward diffusion. Wasserstein \textbf{ distance $W(P^c_t, P^o_t)$ decreases as $t$ increases}. 
\end{proposition}

To solve this problem, we use UOT to map from marginal distribution $q(x_t)$ to backward diffusion $q(x_0|x_t)$, shown in \cref{eq:dduot_obj}. The backward diffusion $q(x_{t-1}|x_t)$ is intractable \cite{ho2020denoising} and it could be written as $q(x_{t-1}|x_t) = \sum_{x_0} q(x_{t-1}|x_t, x_0)q(x_0|x_t)$. From this observation, we formulate the following loss for our framework:

\begin{equation}
    \sum_{t \geq 1}  \UOTsf\left(q\left(x_t\right), q\left(x_{0} \mid x_t \right)\right)
    \label{eq:dduot_obj}
\end{equation}

There are two motivating reasons for using \cref{eq:dduot_obj}. Firstly, since $x_0$ is zero-noised, the distance between outlier and inlier $x_0$ is large and UOT formulation could effectively remove the outliers. This formula helps us avoid the robust ill-posed problem stated by \cref{prop:diffuse}. Secondly, we notice that $q(x_{t-1}|x_t, x_0)$ \cite{ho2020denoising} is tractable and could be easily sampled due to its Gaussian form. Applying the semi-dual UOT \cref{eq12} in the training objective \cref{eq:dduot_obj}, we can obtain:
% \vspace{-5mm}
\begin{align}
\UOTsf(q\left(x_t\right), q\left(x_{0} \mid x_t \right)) 
=&\min _{D_\phi}\bigg[ \Psi_1^*\Big(-\min_{G_\theta}\Big[\tau c\big(x_t, \hat{x}_0\big)-D_\phi\big(\hat{x}_0, x_t, t\big)\Big]\Big) \nonumber\\
&+ \Psi_2^*\big(-D_{\phi}(x_0, x_t, t)\big) \bigg], \label{eq:final}
\end{align}
% \vspace{-5mm}
where  $\hat{x}_0 = G_\theta(x_t, t)$. 

\subsection{Analysis of Semi-Dual UOT formulation}\label{sec:analysis}
In this section, we analyze the importance of choosing $\Psi$ in \cref{eq:final}, the design space of potential network $D_\phi$ and $G_\theta$.

\textbf{Lipschitz property of $\Psi$:} UOTM \cite{choi2023generative} favour the conventional Csisz\'{a}r-divergence $\Psi$ like KL or $\chi^2$. However, in \cref{sc:ablation}, we show that the function, whose convex conjugate is Softplus, performs better than these conventional divergences. As \cite{akbari2021does} states that the Lipschitz loss function results in better performance, we hypothesize that Lipschitz continuity property of Softplus helps the training process more effective while convex conjugate of KL and $\chi^2$ are not Lipschitz (see Appendix 9 for proof of Lipschitz property).

\textbf{Design space of generator function $G_\theta$:}
Motivated from \cite{xiao2022DDGAN}, we also inject latent variable $z \sim \mathcal{N}(0, I)$ as input to $G_\theta$ along with $x_t$ and $t$. There are two reasons for this choice. Firstly, the latent variable $z$ helps the generator mimic stochastic behavior. According to \cite{xiao2022DDGAN}, without latent $z$, the denoising generative model becomes a unimodal distribution, making the sample quality significantly worse. The second reason is that $z$ can be used as style information as in StyleGAN architecture \cite{karras2019style}. Motivated from StyleGAN, DDGAN generator network \cite{xiao2022DDGAN} also uses style modulation layer and AdaIn to inject style information from $z$ into each feature network. As a result, DDGAN inherits the sophisticated architecture of StyleGAN for high-fidelity image synthesis. We adopt a similar architecture design of generator $G_\theta$ from DDGAN \cite{xiao2022DDGAN}.

\textbf{Design space of potential function $D_\phi$:} Through experiment, we discover that using $x_{t-1}$ (instead of $x_0$ in \cref{eq:final}) in potential network $D_\phi$ in place for $x_0$ achieves better FID score. In sampling process, given $x_t$, we predict $\hat{x}_0 = G_\theta(x_t, t, z)$ then draw $x_{t-1} \sim q(x_{t-1}|x_t, \hat{x}_0)$, consequently. The sampling process not only depends on $G_\theta(x_t, t, z)$ but also $q(x_{t-1}|x_t, \hat x_0)$. Therefore, in training framework, we should explicitly use $x_{t-1}$ from $q(x_{t-1}|x_t, \hat x_0)$ as input of potential network to better support the sampling process. Relying on the reason, we propose the modified UOT loss replacing \cref{eq:final}:
\begin{align}
\UOTsf(q\left(x_t\right), q\left(x_{0} \mid x_t \right)) 
=&\min _{D_\phi}\bigg[ \Psi_1^*\Big(-\min_{G_\theta}\Big[\tau c\big(x_t, \hat{x}_0\big)-D_\phi\big(\hat{x}_{t-1}, x_t, t\big)\Big]\Big) \nonumber\\
&+ \Psi_2^*\big(-D_{\phi}(x_{t-1}, x_t, t)\big) \bigg], \label{eq:final2}
\end{align}
where $x_{t-1} \sim q(.|x_t, x_{0})$.

In summary, we present our framework \textbf{\textit{Robust Diffusion Unbalanced Optimal Transport (RDUOT)}} in \cref{alg:SUOT}. In the default setting on clean dataset and outlier robustness, we apply semi-dual UOT to all diffusion steps and use the same cost functions $\mathbf{L}_2$: $c(x, y) = {\tau}||x-y||_2^2$ as UOTM.

% \vspace{-5mm}
\begin{algorithm}[h]
\caption{Robust Diffusion Unbalanced Optimal Transport}\label{alg:SUOT}
    \kwInput{The data distribution $p_{data}$. Non-decreasing, differentiable, a function pair $\left(\Psi_1^*, \Psi_2^*\right)$. Generator network $G_\theta$ and the potential network $D_\phi$. Total training iteration number $K$. Batch size $B$.}\\
    \For{$k=0,1,2, \ldots, K$}{
        Sample $x_0 \sim p_{\text{data}}, z \sim \mathcal{N}(\mathbf{0}, \mathbf{I}_{d}), t\sim [1:T]$.\\
        Sample $x_{t} \sim p(\cdot|x_0), \hat{x}_0 = G_\theta(x_t, z, t), \hat{x}_{t-1} \sim p(\cdot|\hat{x}_0, x_t),{x}_{t-1} \sim p(\cdot|{x}_0, x_t)$.
         \begin{align*}
        \mathcal{L}_D&=\frac{1}{B} \Psi_1^*\big(-c\left(x_t, \hat{x}_0 \right) + D_\phi\left(\hat{x}_{t-1}, x_{t}, t\right)\big) +\frac{1}{B} \Psi_2^*\big(-D_\phi\left(x_{t-1}, x_{t}, t\right)\big)\text {. }
        \end{align*}
        
        Update $\phi$ to minimize the loss $\mathcal{L}_{D}$. % by using the loss $\mathcal{L}_{D}$.
        
        $$
        \mathcal{L}_G=\frac{1}{B} \big(c\left(x_t, \hat{x}_0\right)-D_\phi\left(\hat{x}_{t-1}, x_{t}, t\right)\big) \text {. }
        $$\\
        
        Update $\theta$ to minimize the loss $\mathcal{L}_G$. %by using the loss $\mathcal{L}_G$.
    }

\end{algorithm}
% \vspace{-6mm}

\section{Experiment} \label{sec_experiment}
In this section we firstly show the robustness of our model RDUOT to various corrupted datasets. We then show that RDUOT also possesses high-fidelity generation and fast training convergence properties on clean datasets. Finally, we conduct ablation studies to show the importance of choosing $\Psi$, and to verify the design of our framework  in \cref{sec_method}. Details of all experiments and evaluations can be found in Appendix 7.

\subsection{Robustness to corrupted datasets}
% \begin{table}
% \centering
%   \footnotesize
%     \begin{tabular}{ c | c c | c c}
%         \toprule
%         \multicolumn{1}{c}{} & \multicolumn{2}{|c|}{Synthesized Outlier} & \multicolumn{2}{c}{FID}\\
%         \midrule
%         \multicolumn{1}{c|}{Outlier ratio} & DDGAN & RDUOT & DDGAN & RDUOT \\
%       % Ratio & DDGAN & RDUOT & DDGAN & RDUOT \\
%       \midrule
%       $3\%$ & $3.2\%$ & $0.2\%$  &4.76 &3.43\\
%       $5\%$ & $4.1\%$ & $1.7\%$ &8.81  &4.37\\
%       $7\%$ & $6.9\%$ & $2.3\%$ &9.55  &5.17 \\
%       $10\%$ & $9.8\%$ & $3.8\%$ &14.77  &6.98\\
%       \bottomrule
%     \end{tabular}
%     \captionof{table}{Synthesized Outlier Ratios and FID of DDGAN and RDUOT on datasets (CIFAR10 perturbed by MNIST) with varying outlier ratios.}
%     \label{table:ratio_outlier}
% \end{table}
In this section, we conducted experiments on various datasets perturbed with diverse outlier types, mirroring real-world applications to validate its robustness in handling corrupted datasets. Since the resolution of clean and outlier datasets might be different, we rescaled the clean and outlier datasets to the same resolution, with CI+MI at $32 \times 32$ and the other four datasets (CE+FT, CE+MT, CE+CH and CE+FCE) at $64 \times 64$. Here,   
CI, MI, FT, CE, CH and FCE  stand for CIFAR10, MNIST, FASHION MNIST, CELEBAHQ, LSUN CHURCH and VERTICAL FLIP CELEBAHQ, respectively. "A+B" means "dataset A perturbed with $5\%$ dataset B". 

\minisection{Comparison to DDGAN:}

As shown in \cref{table:ratio_outlier}, our model consistently maintains strong performance even when the outlier percentage in training datasets increases. While the outlier ratio in the training dataset escalates from $3\%$ to $10\%$, RDUOT's FID only increases by around 3.55 points (from $3.43$ to $6.98$). In contrast, DDGAN's FID increases by more than 10 points (from $4.76$ to $14.77$), and the synthesized outlier ratio of RDUOT rises from $0.2\%$ to $3.8\%$ compared to DDGAN's increase from $3.2\%$ to $9.8\%$.

\begin{figure*}[!ht]
% \vspace{-5mm}
    \centering
    \includegraphics[width=0.9\textwidth]{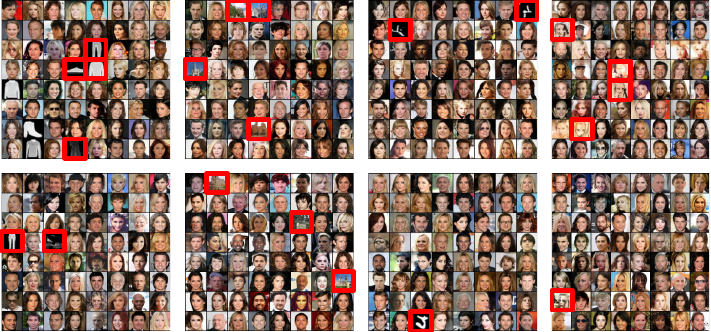}
    \caption{From left to right is corresponding to CE+FT, CE+CH, CE+MT and CE+FCE dataset. Top: DDGAN, Bottom: RDUOT. The red boxes indicate the synthesized outliers among the clean synthesized samples.}
    \label{fig:other_dataset_qual}
% \vspace{-6mm}
\end{figure*}

\begin{table}[ht]
% \vspace{-0.5cm}
  \begin{minipage}{0.6\linewidth}
    \centering 
  \footnotesize
  \begin{tabular}{ c | c c | c c}
        \toprule
        \multicolumn{1}{c}{} & \multicolumn{2}{|c|}{Synthesized Outlier} & \multicolumn{2}{c}{FID}\\
        \midrule
        \multicolumn{1}{c|}{Perturb ratio} & DDGAN & RDUOT & DDGAN & RDUOT \\
      % Ratio & DDGAN & RDUOT & DDGAN & RDUOT \\
      \midrule
      $3\%$ & $3.2\%$ & $0.2\%$  &4.76 &3.43\\
      $5\%$ & $4.1\%$ & $1.7\%$ &8.81  &4.37\\
      $7\%$ & $6.9\%$ & $2.3\%$ &9.55  &5.17 \\
      $10\%$ & $9.8\%$ & $3.8\%$ &14.77  &6.98\\
      \bottomrule
    \end{tabular}
    \caption{Synthesized Outlier Ratios and FID of DDGAN and RDUOT on CIFAR10 (perturbed by MNIST) with varying outlier ratios.}
    \label{table:ratio_outlier}
  \end{minipage}%
  \hfill
  \begin{minipage}{0.35\linewidth}
    \centering 
  \footnotesize
  \begin{tabular}{l|cc}
    \toprule
    & RDUOT & DDGAN\\
    \midrule
    CE+FT  &7.89 &10.68\\
    CE+MT  &9.29 &12.95\\
    CE+CH  &7.86 &9.83\\
    CE+FCE  &5.99 &6.48\\
    \bottomrule
  \end{tabular}
  \caption{FID of DDGAN and RDUOT on CE+FT, CE+CH, CE+MT and CE+FCE.}
  \label{table:ortherdataset}
  \end{minipage}
  % \vspace{-6mm}
\end{table}

% \begin{table}
%   \centering
%   \footnotesize
%   \begin{tabular}{lcc}
%     \toprule
%     & RDUOT & DDGAN\\
%     \midrule
%     CE+FT  &7.89 &10.68\\
%     CE+MT  &9.29 &12.95\\
%     CE+CH  &7.86 &9.83\\
%     CE+FCE  &5.99 &6.48\\
%     \bottomrule
%   \end{tabular}
%   \caption{FID Scores of DDGAN and RDUOT trained on CE+FT, CE+CH, CE+MT and CE+FCE.}
%   \label{table:ortherdataset}
% \end{table}

When testing on higher dimensional datasets, RDUOT keeps dominating DDGAN as can be seen in \cref{table:ortherdataset} and \cref{fig:other_dataset_qual}.
We observe that RDUOT performs well with both outlier datasets FT and MT which are grayscale and visually different from CE, with an FID gap of around 3 points when compared with the corresponding DDGAN model. Notably, even though the CH dataset comprises RGB images and bears great similarity to CE, RDUOT effectively learns to automatically eliminate outliers. For hard outlier dataset FCE, which has a great similarity with CE, RDUOT successfully removes the vertical flip face (refer to last column of \cref{fig:other_dataset_qual}) and we achieve a better FID score compared to DDGAN. This demonstrates RDUOT's capability to discriminate between two datasets in the same RGB domain, which has not previously been explored by other robust generative works \cite{Balaji_Robust, Le_Robust_2021, choi2023generative}.

\minisection{Comparison to other robust frameworks:} As can be seen in \cref{table:orther_robust_algo}, both UOTM \cite{choi2023generative} and RobustGAN \cite{Balaji_Robust} have much higher FID compared to RDUOT. RobustGAN is hard to converge and get very high FID even with two simple corrupted datasets. These results are even worse than DDGAN (\cref{table:ratio_outlier}). For UOTM, we first use KL as $\Psi$, but it cannot learn the data distribution and generate noisy images. We then use Softplus instead and got the FID reported in \cref{table:orther_robust_algo}. However, UOTM still has a lower score compared to RDUOT. Specifically, the FID of UOTM on CE + FCE is higher than DDGAN's FID as shown in \cref{table:ratio_outlier}. These results prove the inferiority of the two existing models compared to RDUOT. 

\begin{table}[]
  \footnotesize
    \centering
    \begin{tabular}{lc|c|c|c|c}
         \toprule
          & CI+3\%MT & CI+5\%MT & CE+FT & CE+CH & CE+FCE\\
          \midrule
          RDUOT & \textbf{3.43}  & \textbf{4.37} & \textbf{7.89} & \textbf{7.86} & \textbf{5.99}\\
         \midrule
          UOTM \cite{choi2023generative} &4.76  &7.89 & 9.52 &8.84 & 6.72\\
         \midrule
          RobustGAN \cite{Balaji_Robust} & 10.63  &10.68 & - & - & - \\
         \bottomrule
    \end{tabular}
    \caption{Robustness comparison on CE+FT, CE+CH, CE+MT and CE+FCE. Note: RobustGAN uses the same architecture as UOTM and RDUOT for fair comparison.}
    \label{table:orther_robust_algo}
\end{table}

\subsection{Performance in clean datasets}
\begin{figure*}[!ht]
% \vspace{-5mm}
\centering
~
\subfloat[CIFAR-10]{
    \includegraphics[width=0.33\linewidth]{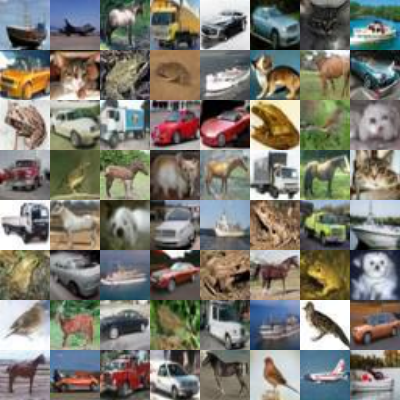}
    \label{fig:cifar10}
}
~
\subfloat[CelebA-HQ]{
    \includegraphics[width=0.33\linewidth]{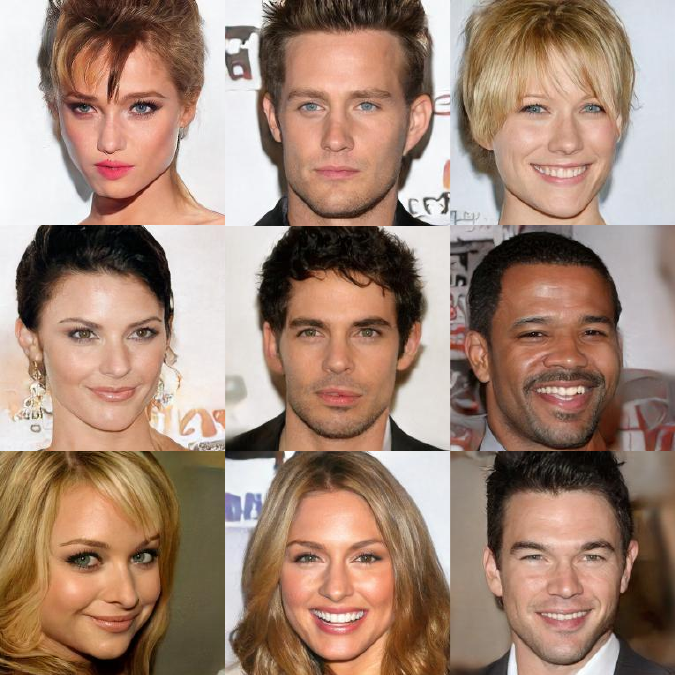}
    \label{fig:celaba-hq}
}
~
\subfloat[STL-10]{
    \includegraphics[width=0.33\linewidth]{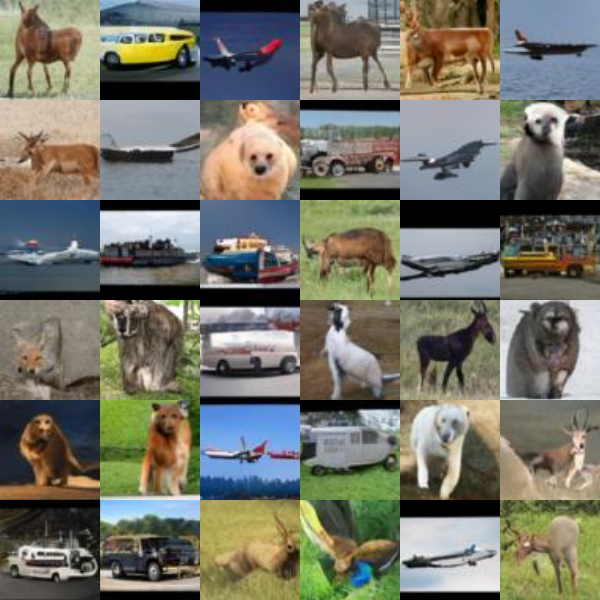}
    \label{fig:stl10}
}
\caption{Qualitative results of RDUOT on 3 datasets STL-10, CIFAR-10, CelebA-HQ.}
    \label{fig:qual_synthesis}
\end{figure*}

We assess the performance of RDUOT technique on three distinct clean datasets: CELEBA-HQ ($256 \times 256$) \cite{karras2017progressive},
CIFAR-10 ($32 \times 32$) \cite{cifar10}, and STL-10 ($64 \times 64$) \cite{stl10} for image synthesis tasks. To assess the effectiveness of RDUOT, we utilize two widely recognized metrics, namely FID \cite{heusel2017gans} and Recall \cite{kynkaanniemi2019improved}. In \cref{tab:FID_clean_cifar} and \cref{tab:quality_cifar10}, we can observe that RDUOT achieves significantly lower FID of $\mathbf{2.95}$ and $\mathbf{5.60}$ for CIFAR10 and CELEBA-HQ, in contrast to the baseline DDGAN, which records FID of $3.75$ and $7.64$ for CIFAR10 and CELEBA-HQ, respectively. Moreover, RDUOT achieves a better Recall of $0.58$ compared to DDGAN's Recall of $0.57$ for CIFAR10 and slightly outperforms DDGAN for CELEBA-HQ with a Recall of $0.38$ compared to DDGAN's $0.36$. 

\begin{table}[ht]
% \vspace{-0.5cm}
  \begin{minipage}{0.45\linewidth}
    \centering 
  \footnotesize
  \begin{tabular}{lccc}
    \toprule
    Model & FID$\downarrow$ & Recall$\uparrow$ & NFE$\downarrow$ \\
    \midrule
    RDUOT &\textbf{2.95} & \textbf{0.58} &4   \\
    WaveDiff \cite{phung2023wavediff} & 4.01 & 0.55 & 4 \\
    DDGAN \cite{xiao2022DDGAN} & 3.75 & 0.57 & 4 \\
    \midrule
    DDPM \cite{ho2020denoising} & 3.21 & 0.57 & 1000\\
    % DDIM \cite{song2020denoising} & 4.67 & 0.53 & 50  \\
    % Recovery EBM \cite{gao2021learning} & 9.58 & - & 180  \\
    \midrule
    StyleGAN2 \cite{karras2020analyzing} & 8.32 & 0.41 & 1 \\
    % NVAE \cite{vahdat2020nvae} & 23.5 & 0.51 & 1  \\
    WGAN-GP \cite{gulrajani2017improved} &39.40 &- &1\\
    RobustGAN \cite{Balaji_Robust} &21.57 &- &1\\
    RobustGAN* &11.40 &- &1\\
    OTM \cite{rout2021generative} &21.78 &- &1\\
    UOTM \cite{choi2023generative} &2.97 &- &1 \\
    $\text{UOTM}^{\#}$ &3.79 &- &1 \\
    \bottomrule
    
  \end{tabular}
\caption{Quantitative results on \\ CIFAR-10. *: DDGAN architecture, $^{\#}$: trained on our machine }
\label{tab:FID_clean_cifar}
  \end{minipage}%
  \hfill
  \begin{minipage}{0.4\linewidth}
    \centering 
  \footnotesize
  \begin{tabular}{l c c }
    \toprule
    Model & FID$\downarrow$ & Recall$\uparrow$  \\
    \midrule
    RDOUT & \textbf{5.60} & \textbf{0.38}  \\
    WaveDiff \cite{phung2023wavediff} & 5.94 & 0.37  \\
    DDGAN \cite{xiao2022DDGAN} & 7.64 & 0.36  \\
    \midrule
    Score SDE \cite{song2020score} & 7.23 & -  \\
    LFM \cite{dao2023flow} &5.26 & - \\
    NVAE \cite{vahdat2020nvae} & 29.7 & -  \\
    VAEBM \cite{xiao2021vaebm} & 20.4 & -  \\
    PGGAN \cite{karras2017progressive} & 8.03 & - \\
    VQ-GAN \cite{esser2020taming} &10.2 & -  \\
    \midrule
    UOTM \cite{choi2023generative} &5.80 &-\\
    \bottomrule
    
    \end{tabular}
\caption{
Quantitative results \\ on CELEBA-HQ. }
\label{tab:quality_cifar10}
  \end{minipage}
  % \vspace{-6mm}
\end{table}

For STL-10 dataset, \cref{tab:stl10} illustrates a substantial improvement in FID for RDUOT compared to DDGAN. Specifically, RDUOT achieves a remarkable FID of $\mathbf{11.50}$, roughly 10 points lower than DDGAN's FID of $21.79$. Additionally, RDUOT achieves a higher Recall of $0.49$, surpassing DDGAN's Recall of $0.40$. Furthermore, RDUOT also outperforms all state-of-the-art methods in terms of FID and Recall.

\begin{table*}[!ht]
% \vspace{-5mm}
\footnotesize
  \begin{varwidth}[b]{0.47\linewidth}
        \raggedright
        \begin{tabular}{l c c }
            \toprule
            Model & FID$\downarrow$ & Recall$\uparrow$ \\
            \midrule
            Our &\textbf{11.50}  & \textbf{0.49} \\
            WaveDiff \cite{phung2023wavediff} & 12.93 & 0.41  \\
            DDGAN \cite{xiao2022DDGAN} & 21.79 & 0.40  \\
            \midrule
            StyleFormer \cite{park2021styleformer} &15.17	&-\\
            TransGAN \cite{jiang2021transgan} &18.28 & -\\
            SNGAN \cite{miyato2018spectral} & 40.1 & -  \\
            StyleGAN2+ADA \cite{karras2020training} & 13.72 & 0.36  \\
            StyleGAN2+Aug\cite{karras2020training} & 12.97 & 0.39 \\
            Diffusion StyleGAN2 \cite{wang2022diffusion} &11.53 & -\\
            \bottomrule
       \end{tabular}
       % \vspace{0.4mm}
       \caption{Quantitative performance of RDUOT on STL-10. RDUOT surpasses DDGAN at both metric FID and Recall.}
       \label{tab:stl10}
  \end{varwidth}%
  \hfill
  \begin{minipage}[b]{0.4\linewidth}
    \centering
    \includegraphics[width=\linewidth]{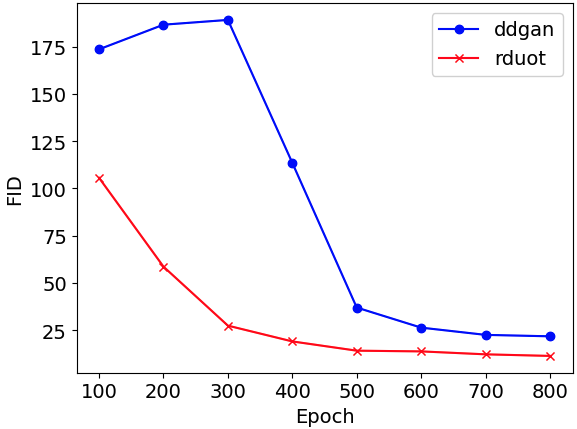}
    \caption{The training convergence on STL-10 between DDGAN and RDUOT.}
    \label{fig:stl10_convergence}
  \end{minipage}
  % \vspace{-6mm}
\end{table*}

In summary, our proposed RDUOT method outperforms the baseline DDGAN in \textbf{high-fidelity image generation} and maintains \textbf{good mode coverage}. In \cref{fig:stl10_convergence}, we demonstrate that RDUOT \textbf{converges much faster} than DDGAN. By epoch $400$, RDUOT achieves an FID of less than $20$, while DDGAN's FID remains above $100$. According to \cite{lou2023reflected}, in training process, stochastic diffusion process can go out of the support boundary, make itself diverge, and thus can generate highly unnatural samples. We hypothesize that the RDUOT's ability to remove outliers at each step (caused by the high variance of large diffusion steps in DDGAN) leads to better performance. For a visual representation of our results, please refer to \cref{fig:qual_synthesis}.

\subsection{Ablation Study} \label{sc:ablation}

\minisection{Selection of $\Psi$:}

Given that $\mathsf{D}_{\Psi_i}$ could be Csisz\'{a}r-divergences, we can choose commonly used functions like KL and $\chi^2$ for $\Psi_1$ and $\Psi_2$ in RDUOT. However, using KL as ${\Psi_i}$ led to infinite loss during RDUOT training, even with meticulous hyperparameter tuning, likely due to the exponential convex conjugate form of KL (refer to Appendix 9). On clean CIFAR-10 dataset, using KL as $\Psi$, we obtain the best FID of 10.11 at epoch 1301 before the loss explodes to $\infty$. This phenomenon shows the instability of KL. For $\chi^2$ as ${\Psi_i}$, the first row of \cref{tab:choice_divergence} reveals that RDUOT with $\chi^2$ achieve a FID score of $5.04$, outperforming DDGAN's FID of $8.81$ on CIFAR-10 with $5\%$ outlier MNIST but still higher than softplus (4.37).

\begin{table}[ht]
  \begin{minipage}{0.5\linewidth}
    \centering 
  \footnotesize
  \begin{tabular}{llcc}
    \toprule
     $\Psi_1^*$ & $\Psi_2^*$ & FID (clean) $\downarrow$ & FID ($5\%$) $\downarrow$ \\
     \midrule
     $\chi^2$   &  $\chi^2$  &3.93  &5.04\\
     \midrule
     softplus   &  softplus  & \textbf{2.95}  & \textbf{4.37}\\
     \bottomrule
  \end{tabular}
\caption{FID for different choices of $\Psi_1^*$ and $\Psi_2^*$.}
\label{tab:choice_divergence}
  \end{minipage}%
  \hfill
  \begin{minipage}{0.4\linewidth}
    \centering 
  \footnotesize
    \begin{tabular}{lcc}
         \toprule
          Outlier ratio &  0\% & 5\%\\
          \midrule % I want to wrap text around this table yes
          Our &2.95 &4.37\\
         \midrule
          Our$^{*}$  &3.09  & 6.94\\
          Our$^{\#}$  &3.94 &5.93\\
         \bottomrule
    \end{tabular}
\caption{Different proposed UOT losses. Our: \cref{eq:final2}, Our$^{*}$: \cref{eq:DDUOT}, Our$^{\#}$: \cref{eq:final}}
\label{table:other_version}
  \end{minipage}
\end{table}

\minisection{Verifying Design of Framework:} \label{sec:ablate_other} 

In this section, we run experiments with other versions of our proposed model for verifying our insight in \cref{sec_robust_diffusion}. The first version uses \cref{eq:DDUOT}, and the second version uses \cref{eq:final}. Their empirical results are shown in \cref{table:other_version}. Since noisy clean and outlier distributions at time $t$ are close to each other, the proposed model using \cref{eq:DDUOT} fails to remove outliers (FID 6.94 compared to 4.37 of the main version). On the other hand, if using \cref{eq:final}, the training process loses the information about $x_{t-1}$ and hurts the sampling process, leading to worse performance as shown in \cref{table:other_version}.

\section{Conclusion}

In this paper, we introduce the first diffusion framework for robust-to-outliers image generation tasks. We present techniques to incorporate UOT into the DDGAN framework, leading to our proposed framework RDUOT. RDUOT has demonstrated the ability to either maintain or enhance performance across all three critical generative modeling criteria: mode coverage, high-fidelity generation, and fast sampling, all while ensuring rapid training convergence. Additionally, our paper showcases that RDUOT significantly outperforms DDGAN and other robust-to-outlier algorithms on corrupted training datasets with various settings, making it a promising approach for real-world corrupted datasets.

% ---- Bibliography ----
%
% BibTeX users should specify bibliography style 'splncs04'.
% References will then be sorted and formatted in the correct style.
%
\bibliographystyle{splncs04}
\bibliography{main}

\begin{thebibliography}{10}
\providecommand{\url}[1]{\texttt{#1}}
\providecommand{\urlprefix}{URL }
\providecommand{\doi}[1]{https://doi.org/#1}

\bibitem{akbari2021does}
Akbari, A., Awais, M., Bashar, M., Kittler, J.: How does loss function affect generalization performance of deep learning? application to human age estimation. In: International Conference on Machine Learning. pp. 141--151. PMLR (2021)

\bibitem{altschuler2017near}
Altschuler, J., Weed, J., Rigollet, P.: Near-linear time approximation algorithms for optimal transport via sinkhorn iteration. In: Advances in Neural Information Processing Systems. pp. 1964--1974 (2017)

\bibitem{arjovsky2017wasserstein}
Arjovsky, M., Chintala, S., Bottou, L.: Wasserstein generative adversarial networks. In: International conference on machine learning. pp. 214--223. PMLR (2017)

\bibitem{Balaji_Robust}
Balaji, Y., Chellappa, R., Feizi, S.: Robust optimal transport with applications in generative modeling and domain adaptation. In: NeurIPS (2020)

\bibitem{chizat2018unbalanced}
Chizat, L., Peyr{\'e}, G., Schmitzer, B., Vialard, F.X.: Unbalanced optimal transport: Dynamic and kantorovich formulations. Journal of Functional Analysis  \textbf{274}(11),  3090--3123 (2018)

\bibitem{choi2023generative}
Choi, J., Choi, J., Kang, M.: Generative modeling through the semi-dual formulation of unbalanced optimal transport. arXiv preprint arXiv:2305.14777  (2023)

\bibitem{stl10}
Coates, A., Ng, A., Lee, H.: An analysis of single-layer networks in unsupervised feature learning. In: Gordon, G., Dunson, D., Dudík, M. (eds.) Proceedings of the Fourteenth International Conference on Artificial Intelligence and Statistics. Proceedings of Machine Learning Research, vol.~15, pp. 215--223. PMLR, Fort Lauderdale, FL, USA (11--13 Apr 2011)

\bibitem{Cuturi-2013-Sinkhorn}
Cuturi, M.: Sinkhorn distances: {L}ightspeed computation of optimal transport. In: Advances in Neural Information Processing Systems. pp. 2292--2300 (2013)

\bibitem{dao2023flow}
Dao, Q., Phung, H., Nguyen, B., Tran, A.: Flow matching in latent space. arXiv preprint arXiv:2307.08698  (2023)

\bibitem{dhariwal2021diffusion}
Dhariwal, P., Nichol, A.: Diffusion models beat gans on image synthesis. Advances in Neural Information Processing Systems  \textbf{34},  8780--8794 (2021)

\bibitem{esser2020taming}
Esser, P., Rombach, R., Ommer, B.: Taming transformers for high-resolution image synthesis (2020)

\bibitem{gallouet2021regularity}
Gallou{\"e}t, T., Ghezzi, R., Vialard, F.X.: Regularity theory and geometry of unbalanced optimal transport. arXiv preprint arXiv:2112.11056  (2021)

\bibitem{gulrajani2017improved}
Gulrajani, I., Ahmed, F., Arjovsky, M., Dumoulin, V., Courville, A.C.: Improved training of wasserstein gans. Advances in neural information processing systems  \textbf{30} (2017)

\bibitem{heusel2017gans}
Heusel, M., Ramsauer, H., Unterthiner, T., Nessler, B., Hochreiter, S.: Gans trained by a two time-scale update rule converge to a local nash equilibrium. Advances in neural information processing systems  \textbf{30} (2017)

\bibitem{ho2020denoising}
Ho, J., Jain, A., Abbeel, P.: Denoising diffusion probabilistic models. In: Advances in neural information processing systems (2020)

\bibitem{huang2021variational}
Huang, C.W., Lim, J.H., Courville, A.C.: A variational perspective on diffusion-based generative models and score matching. Advances in Neural Information Processing Systems  \textbf{34},  22863--22876 (2021)

\bibitem{janati2019spatio}
Janati, H., Cuturi, M., Gramfort, A.: Spatio-temporal alignments: Optimal transport through space and time. arXiv preprint arXiv:1910.03860  (2019)

\bibitem{jiang2021transgan}
Jiang, Y., Chang, S., Wang, Z.: Transgan: Two pure transformers can make one strong gan, and that can scale up. Advances in Neural Information Processing Systems  \textbf{34},  14745--14758 (2021)

\bibitem{karras2017progressive}
Karras, T., Aila, T., Laine, S., Lehtinen, J.: Progressive growing of {GAN}s for improved quality, stability, and variation. In: International Conference on Learning Representations (2018)

\bibitem{karras2020training}
Karras, T., Aittala, M., Hellsten, J., Laine, S., Lehtinen, J., Aila, T.: Training generative adversarial networks with limited data. In: Advances in neural information processing systems (2020)

\bibitem{karras2019style}
Karras, T., Laine, S., Aila, T.: A style-based generator architecture for generative adversarial networks. In: Proceedings of the IEEE/CVF Conference on Computer Vision and Pattern Recognition (2019)

\bibitem{karras2020analyzing}
Karras, T., Laine, S., Aittala, M., Hellsten, J., Lehtinen, J., Aila, T.: Analyzing and improving the image quality of stylegan. In: Proceedings of the IEEE conference on computer vision and pattern recognition (2020)

\bibitem{kingma2021variational}
Kingma, D., Salimans, T., Poole, B., Ho, J.: Variational diffusion models. Advances in neural information processing systems  \textbf{34},  21696--21707 (2021)

\bibitem{kodali2017convergence}
Kodali, N., Abernethy, J., Hays, J., Kira, Z.: On convergence and stability of gans. arXiv preprint arXiv:1705.07215  (2017)

\bibitem{cifar10}
Krizhevsky, A.: Learning multiple layers of features from tiny images. University of Toronto  (05 2012)

\bibitem{kynkaanniemi2019improved}
Kynk{\"a}{\"a}nniemi, T., Karras, T., Laine, S., Lehtinen, J., Aila, T.: Improved precision and recall metric for assessing generative models. Advances in Neural Information Processing Systems  \textbf{32} (2019)

\bibitem{lai1988fenchel}
Lai, H.C., Lin, L.J.: The fenchel-moreau theorem for set functions. Proceedings of the American Mathematical Society  \textbf{103}(1),  85--90 (1988)

\bibitem{Le_Robust_2021}
Le, K., Nguyen, H., Nguyen, Q., Ho, N., Pham, T., Bui, H.: On robust optimal transport: Computational complexity and barycenter computation (2021)

\bibitem{le_etal2023antidreambooth}
Le, T., Phung, H., Nguyen, T., Dao, Q., Tran, N., Tran, A.: Anti-dreambooth: Protecting users from personalized text-to-image synthesis. In: Proceedings of the IEEE/CVF International Conference on Computer Vision (ICCV) (2023)

\bibitem{lou2023reflected}
Lou, A., Ermon, S.: Reflected diffusion models. arXiv preprint arXiv:2304.04740  (2023)

\bibitem{lu2022dpm}
Lu, C., Zhou, Y., Bao, F., Chen, J., Li, C., Zhu, J.: Dpm-solver: A fast ode solver for diffusion probabilistic model sampling in around 10 steps. arXiv preprint arXiv:2206.00927  (2022)

\bibitem{meng2021sdedit}
Meng, C., He, Y., Song, Y., Song, J., Wu, J., Zhu, J.Y., Ermon, S.: Sdedit: Guided image synthesis and editing with stochastic differential equations. arXiv preprint arXiv:2108.01073  (2021)

\bibitem{miyato2018spectral}
Miyato, T., Kataoka, T., Koyama, M., Yoshida, Y.: Spectral normalization for generative adversarial networks. arXiv preprint arXiv:1802.05957  (2018)

\bibitem{park2021styleformer}
Park, J., Kim, Y.: Styleformer: Transformer based generative adversarial networks with style vector (2021)

\bibitem{peyre2019computational}
Peyr{\'e}, G., Cuturi, M.: Computational optimal transport. Foundations and Trends{\textregistered} in Machine Learning  \textbf{11}(5-6),  355--607 (2019)

\bibitem{pham2020-UOT}
Pham, K., Le, K., Ho, N., Pham, T., Bui, H.: On unbalanced optimal transport: An analysis of sinkhorn algorithm (2020)

\bibitem{phung2023wavediff}
Phung, H., Dao, Q., Tran, A.: Wavelet diffusion models are fast and scalable image generators. In: Proceedings of the IEEE/CVF Conference on Computer Vision and Pattern Recognition (CVPR). pp. 10199--10208 (June 2023)

\bibitem{ramesh2022hierarchical}
Ramesh, A., Dhariwal, P., Nichol, A., Chu, C., Chen, M.: Hierarchical text-conditional image generation with clip latents. arXiv preprint arXiv:2204.06125  (2022)

\bibitem{rombach2022high}
Rombach, R., Blattmann, A., Lorenz, D., Esser, P., Ommer, B.: High-resolution image synthesis with latent diffusion models. In: Proceedings of the IEEE/CVF Conference on Computer Vision and Pattern Recognition. pp. 10684--10695 (2022)

\bibitem{rout2021generative}
Rout, L., Korotin, A., Burnaev, E.: Generative modeling with optimal transport maps. arXiv preprint arXiv:2110.02999  (2021)

\bibitem{ruiz2022dreambooth}
Ruiz, N., Li, Y., Jampani, V., Pritch, Y., Rubinstein, M., Aberman, K.: Dreambooth: Fine tuning text-to-image diffusion models for subject-driven generation  (2022)

\bibitem{saharia2022photorealistic}
Saharia, C., Chan, W., Saxena, S., Li, L., Whang, J., Denton, E., Ghasemipour, S.K.S., Ayan, B.K., Mahdavi, S.S., Lopes, R.G., et~al.: Photorealistic text-to-image diffusion models with deep language understanding. arXiv preprint arXiv:2205.11487  (2022)

\bibitem{salimans2018improving}
Salimans, T., Zhang, H., Radford, A., Metaxas, D.: Improving gans using optimal transport. arXiv preprint arXiv:1803.05573  (2018)

\bibitem{sanjabi2018convergence}
Sanjabi, M., Ba, J., Razaviyayn, M., Lee, J.D.: On the convergence and robustness of training gans with regularized optimal transport. Advances in Neural Information Processing Systems  \textbf{31} (2018)

\bibitem{sohl2015deep}
Sohl-Dickstein, J., Weiss, E., Maheswaranathan, N., Ganguli, S.: Deep unsupervised learning using nonequilibrium thermodynamics. In: International Conference on Machine Learning (2015)

\bibitem{song2020denoising}
Song, J., Meng, C., Ermon, S.: Denoising diffusion implicit models. In: International Conference on Learning Representations (2021)

\bibitem{song2021maximum}
Song, Y., Durkan, C., Murray, I., Ermon, S.: Maximum likelihood training of score-based diffusion models. Advances in Neural Information Processing Systems  \textbf{34},  1415--1428 (2021)

\bibitem{song2019generative}
Song, Y., Ermon, S.: Generative modeling by estimating gradients of the data distribution. In: Advances in neural information processing systems (2019)

\bibitem{song2020score}
Song, Y., Sohl-Dickstein, J., Kingma, D.P., Kumar, A., Ermon, S., Poole, B.: Score-based generative modeling through stochastic differential equations. In: International Conference on Learning Representations (2021)

\bibitem{vacher2022stability}
Vacher, A., Vialard, F.X.: Stability and upper bounds for statistical estimation of unbalanced transport potentials. arXiv preprint arXiv:2203.09143  (2022)

\bibitem{vahdat2020nvae}
Vahdat, A., Kautz, J.: {NVAE}: A deep hierarchical variational autoencoder. In: Advances in neural information processing systems (2020)

\bibitem{Villani2008OptimalTO}
Villani, C.: Optimal transport: Old and new (2008)

\bibitem{wang2022semantic}
Wang, W., Bao, J., Zhou, W., Chen, D., Chen, D., Yuan, L., Li, H.: Semantic image synthesis via diffusion models. arXiv preprint arXiv:2207.00050  (2022)

\bibitem{wang2022diffusion}
Wang, Z., Zheng, H., He, P., Chen, W., Zhou, M.: Diffusion-gan: Training gans with diffusion. arXiv preprint arXiv:2206.02262  (2022)

\bibitem{xiao2021vaebm}
Xiao, Z., Kreis, K., Kautz, J., Vahdat, A.: Vaebm: A symbiosis between variational autoencoders and energy-based models. In: International Conference on Learning Representations (2021)

\bibitem{xiao2022DDGAN}
Xiao, Z., Kreis, K., Vahdat, A.: Tackling the generative learning trilemma with denoising diffusion {GAN}s. In: International Conference on Learning Representations (ICLR) (2022)

\bibitem{yang2018scalable}
Yang, K.D., Uhler, C.: Scalable unbalanced optimal transport using generative adversarial networks. arXiv preprint arXiv:1810.11447  (2018)

\bibitem{zhang2022fast}
Zhang, Q., Chen, Y.: Fast sampling of diffusion models with exponential integrator. arXiv preprint arXiv:2204.13902  (2022)

\end{thebibliography}

\clearpage
% \setcounter{page}{1}
% \maketitlesupplementary

\title{Supplementary Material}

% \maketitle

% \begin{abstract}

%  In this supplementary material, we firstly provide implementation details are also given for better reproducibility purpose. Secondly, we derive proof for \cref{prop:diffuse} in the main paper. Thirdly, we explain the criteria for choosing $\Psi$ in detail. We then present several additional results on toy dataset and how to apply other noisy preprocessing with our robustness technique in parallel. Furthermore, we also conduct experiment on RDUOT with different number of timesteps. Our code is also included in the supplementary package. 
% \keywords{Diffusion Model \and Unbalanced Optimal Transport \and Robustness Generation \and OT-based Generative Model}
% \end{abstract}

% % \section{Rationale}
% % \label{sec:rationale}

\section{Detailed Experiments} \label{appen:A}

\subsection{Network configurations}

\subsubsection{Generator.} Our generator follows a UNet-like architecture primarily inspired by NCSN++ \cite{song2020score, xiao2022DDGAN}. Detailed configurations of the generator for each dataset can be found in \cref{tab:net_config}.

\subsubsection{Discriminator.} The discriminator has the same number of layers as the generator. Further details about the discriminator's structure can be found in \cite{xiao2022DDGAN}.

\begin{table*}[!ht]
\centering
\small
\begin{tabular}{|l|c|c|c|c|c|c|}
\toprule
                                  & CIFAR & STL &CELEBA    &CI+MT  & CE+\{CH,FT,MT\} & CE+FCE\\
\midrule
% \# of timesteps                   & 4        &4      &2    &4         &2 &2\\
\# of ResNet/scale     & 2        & 2    &2     & 2        &2 &2  \\
Base channels                     & 128      & 128   &64    &128       &96  &128\\
Ch mult/scale & 1,2,2,2  & 1,2,2,2 &1,1,2,2,4,4 &1,2,2,2     & 1,2,2,2,4 & 1,2,2,2,4\\
Attn resolutions             & 16        & 16   &16    & 16      & 16 & 16 \\
Latent Dimension                  & 100      & 100    &100   & 100     & 100 & 100 \\
\#'s latent mapping       & 4        & 4      &4   & 4       & 4 & 4 \\
Latent-embed dim        & 256      & 256   &256    & 256     & 256 & 256 \\
\bottomrule
\end{tabular}
\caption{Network configurations. }
\label{tab:net_config}
\end{table*}

\subsection{Training Hyperparameters}
For the sake of reproducibility, we have provided a comprehensive table of tuned hyperparameters in \cref{tab:hyperparams}. Our hyperparameters align with the baseline \cite{xiao2022DDGAN}, with minor adjustments made only to the number of epochs and the allocation of GPUs for specific datasets. In terms of training times, models for CIFAR-10 and STL-10 take 1.6 and 3.6 days, respectively, on a single GPU. For CI-MT and CE+\{CH,FT,MT\}, it takes 1.6 and 2 day GPU hours, correspondingly.

\begin{table*}[!ht]
    \centering
    \small
    \begin{tabular}{|l|c|c|c|c|c|c|}
        \toprule
                                            & CIFAR-10 & STL-10  &CELEBAHQ  &CI+MT  & CE+\{CH,FT,MT\} & CE+FCE\\
        \midrule
        $\text{lr}_{G}$                            & $1.6\text{e-4}$  & $1.6\text{e-}4$ &$1\text{e-}4$ &$1.6\text{e-}4$      & $1.6\text{e-}4$   & $8\text{e-}5$\\
        $\text{lr}_{D}$                            & $1.25\text{e-}4$ & $1.25\text{e-}4$ &$2\text{e-}4$ & $1.25\text{e-}4$      & $1.\text{e-}4$ & $5\text{e-}5$    \\
        Adam $\beta_1$\&$\beta_2$ & 0.5, 0.9  & 0.5, 0.9 & 0.5, 0.9 & 0.5, 0.9 & 0.5, 0.9  & 0.5, 0.9 \\
        EMA                                 & 0.9999    & 0.9999  & 0.9999  & 0.999       & 0.999  & 0.9999    \\
        Batch size                          & 256       & 72       & 12   &48    & 72  & 96   \\
        Lazy regularizer                 & 15        & 15    &10    & 15          & 15     & 15  \\
        \# of epochs                        & 1800      & 1200  &800     & 1800     & 800   & 800  \\
        \# of timesteps                     & 4         & 4     &2    & 4           & 2        & 2    \\
        \# of GPUs                          & 1         & 1     &2    & 2           & 1  & 2  \\
        r1 gamma                        &0.02        &0.02    &2.0     &0.02      &0.02   &2\\
        Tau $\tau$ for our                          &1e-3    &1e-4  &1e-7  &1e-3  &3e-4 &1e-3\\
        \bottomrule
    \end{tabular}
    \caption{Choices of hyper-parameters}
    \label{tab:hyperparams}
\end{table*}

\subsection{Dataset Preparation}

\minisection{Clean Dataset:} We conducted experiments on two clean datasets CIFAR-10 ($32 \times 32$) and STL-10 ($64 \times 64$). For training, we use 50,000 images.

\minisection{Noisy Dataset:} 

\begin{itemize}
    \item \textbf{CI+MI}: We resize the MNIST data to resolution of ($32\times32$) and mix into CIFAR-10. The total samples of this dataset is 50,000 images.
    \item \textbf{CE+\{CH,FT,MT,FCE\}}: We resize the CelebHQ and CIFAR-10, FASHION MNIST, LSUN CHURCH to the resolution of ($64\times64$), flip CelebHQ vertically, and mix them together. The CelebHQ is clean, and the others are outlier datasets. The noisy datasets contain 27,000 training images.
\end{itemize}

\subsection{Evaluation Protocol}
We measure image fidelity by Frechet inception distance (FID)~\cite{heusel2017gans} and measure sample diversity by Recall metric~\cite{kynkaanniemi2019improved}.

\minisection{FID:} We compute FID between ground truth clean dataset and 50,000 generated images from the models

\minisection{Recall:} Similar to FID, we compute Recall between ground truth dataset and 50,000 generated images from the models

\minisection{Outlier Ratio:} We train classifier models between clean and noisy datasets and use them to classify the synthesized outliers. We first generate 50,0000 synthesized images and count all the outliers from them.

% \section{Alternative loss function}
% In \cref{eq:final}, the cost function can be replaced by the distance between $x_t$ and $\hat{x}_{t-1}$, where $\widehat{x}_{t-1} \sim q(x_{t-1}|x_t,\widehat{x}_0)$ with $\widehat{x}_0 = G_\theta(x_t, z, t)$ ($z \sim \mathcal{N}(0, \mathbb{I})$). We have alternative objective equation:

% \begin{align}
% &\UOTsf(q\left(\mathbf{x}_{t-1} \mid \mathbf{x}_t\right) , p_\theta\left(\mathbf{x}_{t-1} \mid \mathbf{x}_t\right)) \nonumber\\
% =&\min _{D_\phi}\bigg[ \Psi_1^*\Big(-\min_{G_\theta}\Big[c\big(x_t, \widehat{x}_t\big)-D_\phi\big(\widehat{x}_{t-1}, x_t, t\big)\Big]\Big) \nonumber\\
% &+ \Psi_2^*\big(-D_{\phi}(x_{t-1}, x_t, t)\big) \bigg]. \label{eq:final_alternative}
% \end{align}

% \section{Modification in RDUOT with a view from DDGAN and UOTM}

\section{Proofs}

%\begin{proposition}
% \label{tm:diffuse}
%Let $P$ and $Q$ be two probability measures in the set of probability measures, two random variables $X \sim P$ and $Y \sim Q$, $0 \leq \alpha < 1$. We have $W(\mathcal{N}(\sqrt{\alpha}X, (1 - \alpha)\mathbf{I}), \mathcal{N}(\sqrt{\alpha}Y, (1 - \alpha)\mathbf{I})) \leq \sqrt{\alpha} W(P, Q)$, with $W(\cdot, \cdot)$ be the Wasserstein-2 distance.
%\end{proposition}
\begin{proposition}[Restated] \label{proposition_1_appendix}
Denote $P^c$ and $P^o$ be   probability measures of clean outlier data. Define $P_t$ be the probability measure such that $x_t \sim P_t$ is obtained from $x_0 \sim P$ by a forward diffusion process. Then the Wasserstein distance $W(P^c_t, P^o_t)$ decreases as $t$ increases, where $P^c_t$ and $P^o_t$ are probability measures of clean and outlier data after $t$ steps of rescaling and adding noise. 
\end{proposition}
Before we present the proof of Proposition \ref{proposition_1_appendix}, we need the following results in Lemma \ref{lemma_wasserstein}. For simplicity, given any random variable $X$, we denote $q_X$ to be the distribution  function of $X$. 
\begin{lemma}\label{lemma_wasserstein}
    Let $X, Y, Z_1, Z_2$ be random vectors in $\mathbb{R}^d$ such that and $Z_1$ and $Z_2$ are i.i.d and both are independent with $X$ and $Y$. Let $\beta$ be a constant in $(0,1)$. Let $q_{X}, q_Y, q_{\beta X}, q_{\beta Y}, q_{X+Z_1}, q_{Y+Z_2}$ be the distribution functions of $X, Y,\beta X, \beta Y, X+Z_1, X+Z_2$, respectively. Then
    \begin{align}
        W_2(q_{\beta X}, q_{\beta Y}) &\leq \beta W_2(q_X, q_Y)
        \label{lemma_scale}\\
        W_2(q_{X + Z_1}, q_{Y + Z_2}) &\leq W_2(q_X, q_Y).
        \label{lemma_translation}
    \end{align}
    where $W_2$ is the Wasserstein 2-distance between two distributions.
\end{lemma}
% To prove \cref{lemma_scale}, we just need to point out a transport map form $q(\alpha x)$ to $q(\alpha y)$ that has cost $\alpha W(q(x), q(y))$.
\begin{proof} 
Let $x =  (x_1,\ldots,x_d)$ and $ t= (t_1,\ldots,t_d)\in \mathbb{R}^d$, we use notation $\prec$ in  $x\prec t$ means $x_i \leq t_i$, for all $1\leq i\leq d$.  By definition of $q_X$, we have
\begin{align*}
    \mathbb{P}\big(X\prec t \big) = \int_{x\prec t} q_X(x) dx.
\end{align*}
It follows that
\begin{align*}
    \mathbb{P}(\beta X\prec t) = \mathbb{P}\big(X\prec \frac{1}{\beta} t\big) =\int_{x\prec \frac{1}{\beta} t} q_X(x) dx.
\end{align*}
Taking derivative with respect to $t$, we get
\begin{align*}
    q_{\beta X}(t) = \frac{1}{\beta}q_{X}\big(\frac{t}{\beta}\big).
\end{align*}
Let $q_{X,Y}^*$ be the optimal transport density between $q_X$ and $q_Y$. Scale the source and target points in $q_{X,Y}^*$ by a factor of $\beta$, we define a transport plan 
\begin{align*}
q_{\beta X,\beta Y}^*(x,y) = \frac{1}{\beta^2} q_{X,Y}^*(\frac{x}{\beta},\frac{y}{\beta})
\end{align*}
between $q_{\beta X}$ and $q_{\beta Y}$. We verify it by checking its marginal distributions,
\begin{align*}
    \int_{y}q_{\beta X,\beta Y}^*(x,y) dy &= \frac{1}{\beta^2}\int_y q_{X,Y}^*\big(\frac{x}{\beta},\frac{y}{\beta} \big) dy = \frac{1}{\beta}\int_{\frac{y}{\beta}} q_{X,Y}^*\big(\frac{x}{\beta},\frac{y}{\beta} \big) d\frac{y}{\beta} \\
    &= \frac{1}{\beta} q_X\big(\frac{x}{\beta} \big) = q_{\beta X}(x).
\end{align*}
Hence, the marginal distributions of $q_{\beta X,\beta Y}^*$ are $q_{\beta X}$ and $q_{\beta Y}$. Furthermore,
\begin{align*}
   W_2^2( q_{\beta X},q_{\beta Y}) &\leq  \int_{x,y} \|x-y\|^2 q_{\beta X,\beta Y}^*(x,y) dx dy  \\
   &=  \int_{x,y} \|x-y\|^2 \frac{1}{\beta^2} q_{X,Y}^*\big(\frac{x}{\beta},\frac{y}{\beta}\big)  dx dy\\
    &= \int_{x^{\prime},y^{\prime}} \beta^2 \|x^{\prime}-y^{\prime}\|^2 q_{X,Y}^*(x^{\prime},y^{\prime})dx^{\prime} dy^{\prime} \\
    &= \beta^2 W_2^2(q_X,q_Y)
\end{align*}
where $x^{\prime} = \frac{x}{\beta}, y^{\prime} =  \frac{y}{\beta}$. Taking the square root of both sides, we obtain the first inequality.

%Since $W(\cdot, \cdot)$ is the Wasserstein-2 distance, the cost of $\pi_\beta$ is $\beta W(q_X, q_Y)$. Then, the optimal transport cost between $q_{\beta X}$ and $q_{\beta Y}$ can not be bigger than $\beta W(q_X, q_Y)$ (\cref{lemma_scale} is proved).

Let $q_Z$ be the distribution function of $Z_1$ and $Z_2$. Given $q_{X,Y}^*$ is the optimal transport map from $q_X$ to $q_Y$, we again are going to build transport density from $q_{X+Z_1}$ to $q_{Y+Z_2}$ as follow
\begin{align*}
    q^*(x_1, y_1) = \int_tq_{X,Y}^*(x_1-t,y_1-t) q_Z(t) dt.
\end{align*}
First we find the marginal distributions of $q^*$ 
\begin{align*}
    \int_{y_1}q^*(x_1,y_1) dy_1  &= \int_{y_1} \int_t q_{X,Y}^*(x_1-t,y_1-t) q_Z(t) dt dy_1 \\
    &= \int_t\int_{y_1} q^*_{X,Y}(x_1-t,y_1-t) dy_1 q_Z(t)dt \\
    &= \int_t \int_{s} q_{X,Y}^*(x_1-t,s) ds q_Z(t) dt \\
    &= \int_t q_{X}(x_1-t) q_Z(t) dt = q_{X+Z_1}(x_1),
\end{align*}
where $s = y_1 -t$.
Similarly, we have
\begin{align*}
    \int_{x_1}q^*(x_1,y_1) d x_1 = q_{Y+Z_2}(y_1).
\end{align*}
Hence, $q^*$ has marginal distributions  $q_{X+Z_1}$ and $q_{Y+Z_2}$. We next prove the second inequality,
\begin{align*}
    &W_2^2(q_{X+Z_1},q_{Y+Z_2})\\
    &\leq \int_{x_1,y_1} \|x_1 - y_1\|^2 q^*(x_1,y_1) dx_1 dy_1 \\
    &= \int_{x_1,y_1}\int_t \|x_1 - y_1\|^2 q^*_{X,Y}(x_1-t,y_1-t)q_Z(t) dt dx_1 dy_1  \\
    &= \int_t \int_{x_1,y_1}\|(x_1-t) - (y_1-t)\|^2 q^*_{X,Y}(x_1-t,y_1-t) dx_1 dy_1 q_Z(t) dt \\
    &=\int_t \int_{u,v}\|u-v\|^2 q^*_{X,Y}(u,v) du  dv   q_Z(t) dt \\
    &= W_2^2(q_X,q_Y) \int_tq_Z(t) dt = W_2^2(q_X,q_Y),
\end{align*}
where $u = x_1-t$ and $v = y_1-t$.
\end{proof}

%Since adding the same constants to two random variables will not change the Wasserstein distance between them, then we have:

%$$W(q_{X + Z | Z = z_0}, q_{Y + Z| Z = z_0}) = W(q_X, q_Y)$$

%As a result:

%\begin{align}
%\int_{z \sim q_Z} W^2 (q_{X + Z | Z = z}, q_{Y + Z | Z = z}) q_Z(z) dz 
%&= \int_{z \sim q_Z} W^2 (q_X, q_Y) q_Z(z) dz\\
%&= W^2 (q_X, q_Y)
%\end{align}

%Then, there is a transport map between $q_{X + Z}$ and $q_{Y + Z}$ whose cost is $W(q_X, q_Y)$ (\cref{lemma_translation} is proved).

\begin{proof}[\ref{proposition_1_appendix}]
Recall that 
\begin{align*}
    x_t^o = \sqrt{\alpha_t} x_{t-1}^o + \sqrt{1-\alpha_t} Z_{1t}; \quad Z_{1t} \sim N\big(\mathbf{0},\mathbf{I}\big) \\
     x_t^c = \sqrt{\alpha_t} x_{t-1}^c + \sqrt{1-\alpha_t} Z_{2t}; \quad Z_{2t} \sim N\big(\mathbf{0},\mathbf{I}\big)
\end{align*}
where $Z_{1t}$ and $Z_{2t}$ are independent. Let
$q_{X_t^o}$, $q_{X_t^c}$, $q_{X_{t-1}^o}$ and $q_{X_{t-1}^c}$ be the distribution functions of $x_t^o$, $x_{t}^c$, $x_{t-1}^o$ and $x_{t-1}^c$, respectively. By Lemma \ref{lemma_wasserstein}, we have
\begin{align*}
    W_2\big(q_{X_t^o},q_{X_t^c}\big) \leq \sqrt{\alpha_t} W_2\big(q_{X_{t-1}^o},q_{X_{t-1}^c}\big) < W_2\big(q_{X_{t-1}^o},q_{X_{t-1}^c}\big).
\end{align*}

%Let $U = \sqrt{\alpha} X + \sqrt{1 - \alpha} I$, $V = \sqrt{\alpha} Y + \sqrt{1 - \alpha} I$ ($I \sim \mathcal{N}(0, I)$). Then, $U \sim \mathcal{N}(\sqrt{\alpha}X, (1 - \alpha)\mathbf{I})$ and $V \sim \mathcal{N}(\sqrt{\alpha}Y, (1 - \alpha)\mathbf{I})$

%Using \cref{lemma_wasserstein}, we have:

%$$W(q_{\sqrt{\alpha} X + \sqrt{1 - \alpha} I}, q_{\sqrt{\alpha} Y + \sqrt{1 - \alpha} I}) \leq \sqrt{\alpha}W(q_X, q_Y)$$ 

%Or:

%$$W(\mathcal{N}(\sqrt{\alpha}X, (1 - \alpha)\mathbf{I}), \mathcal{N}(\sqrt{\alpha}Y, (1 - \alpha)\mathbf{I})) \leq \sqrt{\alpha} W(P, Q)$$
\end{proof}
\cref{proposition_1_appendix} is a consequence of \cref{lemma_wasserstein}.

\section{Criteria for choosing $\Psi$} \label{appen:B}

To choose $\Psi_1$ and $\Psi_2$ for this loss function, we recommend two following criteria. First, $\Psi_1$ and $\Psi_2$ could make the trade-off between the transport map $\pi$ in \cref{formulation:UOT}  and the hard constraint on the marginal distribution $\mu$ to $\nu$ in order to seek another relaxed plan that transports masses between their approximation but may sharply lower the transport cost. From the view of robustness, this relaxed plan can ignore some masses from the source distribution of which the transport cost is too high, which can be seen as outliers. Second, $\Psi_1$ and $\Psi_2$ need to be convex and differentiable so that  \cref{eq_uot_semi} holds.

Two commonly used candidates for $\Psi_1$ and $\Psi_2$ are two $f$-divergences 
\begin{align*}
\text{KL divergence:} \quad f(x) &= \begin{cases}
x \ln x &\quad x > 0\\
\infty &\quad \text{otherwise}
\end{cases}\\
\chi^2: \quad f(x) &= \begin{cases}
(x-1)^2 &\quad x >0 \\
\infty &\quad \text{otherwise}
\end{cases}
\end{align*}
%KL ($f(x) = x \ln x$ if $x > 0$ else $f(x) = \infty$) and  $\chi^2$ ($f(x) = (x-1)^2$ if $x > 0$ else $f(x) = \infty$)
% Two commonly used candidates for $\Psi_1$ and $\Psi_2$ are two f-divergence KL ($f(x) = x \ln x$ or $f(x) = x \ln x - x + 1$ \cite{choi2023generative} for $x > 0$ else $f(x) = \infty$) and  $\chi^2$ ($f(x) = (x-1)^2$ if $x > 0$ else $f(x) = \infty$). 
However, the convex conjugate of KL is an exponential function, making the training process for DDGAN complicated due to the dynamic of loss value between its many denoising diffusion time steps. Among the ways we tune the model, the loss functions of both generator and discriminator models keep reaching infinity.

Thus, we want a more "stable" convex conjugate function. That of $\chi^2$ is quadratic polynomial, which does not explode when $x$ increases like that of KL:
\begin{equation}
\label{eq:chi_square_convex_conjugate}
    \Psi^*(x)= \begin{cases}\frac{1}{4} x^2+x, & \text { if } x \geq-2 \\ -1, & \text { if } x<-2\end{cases}
\end{equation}
But it is still not Lipschitz continuous. 

As stated in section \cref{sec_method}, we hypothesize that Lipschitz continuity of Softplus can raise the training effectiveness while convex conjugate of KL and $\chi^2$ are not Lipschitz.
Here, we provide the proofs of Lipschitz.
But first, we reiterate that the convex conjugate of a function $f: \mathbb{R} \rightarrow (-\infty, \infty)$ is defined as:
\begin{equation}
\label{eq:convex_conjugate}
    f^*(x) = \sup_{y \in \mathbb{R}} \{\langle x, y \rangle - f(y)\}.
\end{equation}
\textbf{a) Convex conjugate of KL function is non-Lipschitz:}
We have
\begin{align*}
    \Psi(y) = \begin{cases} y \ln y &\quad y > 0\\
    \infty &\quad \text{otherwise}.
    \end{cases}
\end{align*}
%Let $\Psi(y) = y \ln y$ if $y > 0$ else $\Psi(y) = \infty$ 
Thus
\begin{equation}
    \Psi^*(x) = e^{x - 1}.
\end{equation}
% We will prove the statement by contradiction. Supposed $\Psi$ is Lipschitz continuous on $\mathbb{R}$ there exists a Lipschitz constant $L$. 
Choose $x_1 = a + 1$ and $x_2 = a + 1 + \epsilon$, $\epsilon > 0$. We have:
\begin{align}
    \frac{|\Psi^*(x_1) - \Psi^*(x_2)|}{|x_2 - x_1|} 
      = \frac{|e ^ {a + \epsilon} - e^a|}{\epsilon} 
      = e^a \frac{|e^{\epsilon} - 1|}{\epsilon}
      \overset{a \to \infty}{\longrightarrow} \infty
\end{align}
Thus, $\frac{|\Psi^*(x_2) - \Psi^*(x_2)|}{|x_2 - x_1|}$ does not have an upper bound, and convex conjugate of KL function is non-Lipschitz.

\noindent \textbf{b) Convex conjugate of $\chi^2$ function is non-Lipschitz:} Convex conjugate of $\chi^2$ function $\Psi*$ is defined as \cref{eq:chi_square_convex_conjugate}.
% We will prove the statement by contradiction. Supposed $\Psi$ is Lipschitz continuous on $\mathbb{R}$ there exists a Lipschitz constant $L$.
Choose $x_1 = a$ and $x_2 = a + \epsilon$, $a > 0$, $\epsilon > 0$. We have:
\begin{align}
    \frac{|\Psi^*(x_2) - \Psi^*(x_2)|}{|x_2 - x_1|} 
      &= \frac{|0.5 a \epsilon + 0.25 \epsilon ^ 2 + \epsilon|}{\epsilon}= |0.5 a + 0.25 \epsilon + 1| \\
      &\overset{a \to \infty}{\longrightarrow} \infty
\end{align}
Thus, $\frac{|\Psi^*(x_2) - \Psi^*(x_2)|}{|x_2 - x_1|}$ does not have an upper bound, and convex conjugate of KL function is non-Lipschitz.

\noindent \textbf{c) Softplus has Lipschitz continuity property:}

\noindent We have  $\Psi^*(x) = \ln(1 + e ^x)$, $a > 0$. Then
\begin{align}
    |\Psi^*(x + a) - \Psi^*(x)| &= \big|\ln(1 + e ^ {x + a}) - \ln(1 + e^x)\big| = \Big|\ln(\frac{1 + e ^ {x + a}}{1 + e^x})\Big| \\
      &\leq |\ln(\frac{e^a + e ^ {x + a}}{1 + e^x})| = |\ln(e^a)| = a < 2a \\
      &= 2 |(x + a) - x| 
\end{align}

% This is the reason we try $\Psi^*$ as softplus in the loss function, and it turns out to be an interestingly useful discovery.

\minisection{Remark:} For any function $f$, its convex conjugate is always semi-continuous, and $f=f^{**}$ if and only if $f$ is convex and lower semi-continuous \cite{lai1988fenchel}. So, we can choose $f^*$ first such that this is a non-decreasing, differentiable, and semi-continuous function. Then, we find $f^{**}$ and check if $f^{***}$ and $f^*$ is equal. If $f^{***}$ and $f^*$, $f^{**}$ will be a function of which convex conjugate is $f^{*}$. Then we will check if $f^{**}$ satisfied the first criterion to use it as $\Psi_1$ or $\Psi_2$.

% With this remark, we can give proof of proposition \cref{prop:linear_uotm}.

With this remark, we can see why \textbf{functions whose convex conjugate is a simple linear function cannot filter out outliers}.

If $\Psi^*(x) =  ax + b$, $(a > 0)$, we have:

\begin{equation}
\label{eq:linear_psi}
    \Psi(x)= \begin{cases}
         a, & \text { if } x = a \\ 
        \infty, & \text { if } x \neq a
    \end{cases}
\end{equation}

As a result, with equation \cref{formulation:UOT}, the UOTM cost is finite only when $\frac{d \pi_1}{d \mu} = \frac{d \pi_2}{d \nu} = a$ (constant). We will prove the unbalanced optimal transport map is the same as the optimal transport map of the origin OT problem scaled by $a$.

Let $\pi^*$ be the optimal transport map of the OT problem \cref{formulation:OT_sup}. Then, the marginal distribution of $\pi^*$ is $\mu$ and $\nu$. Recall that
\begin{align} 
    \OTsf(\mu,\nu) &= \min_{\pi \in \Pi(\mu,\nu)}\int c(x,y) d\pi(x,y), \label{formulation:OT_sup}\\
    \UOTsf(\mu,\nu) &= \min_{\pi \in \gM(\gX\times \gX)}\int c(x,y) d\pi(x,y) +  \mathsf{D}_{\Psi}(\pi_1\|\mu) + \mathsf{D}_{\Psi}(\pi_2\|\nu), \label{formulation:UOT_sup}
\end{align}

% Let $\bar{\pi} = a \pi^*$, the marginal distribution of $\bar{\pi} $ is $\bar{\pi}_1 = a \mu$ and $\bar{\pi}_2 = a \nu$, which satisfies $\frac{d \bar{\pi}_1}{d \mu} = \frac{d \bar{\pi}_2}{d \nu} = a$.
% Supposed $\bar{\pi}$ is not the optimal transport map of \cref{formulation:UOT_sup} but $\hat{\pi}$. Then, $\int c(x,y) d \hat{\pi} < \int c(x,y) d \bar{\pi} $, and the marginal distribution of $\hat{\pi}$ also satisfies $\frac{d \hat{\pi}_1}{d \mu} = \frac{d \hat{\pi}_2}{d \nu} = a$. As a result, $\frac{\hat{\pi}} {a}$ is the optimal transport map of \cref{formulation:OT_sup}; which means $\int c(x,y) d \frac{\hat{\pi}}{a} = \int c(x,y) d \pi^*$.

%\begin{align} 
 %   \UOTsf(\mu,\nu) &= \min_{\pi \in \gM(\gX\times \gX)}\int c(x,y) d\pi(x,y) +  \mathsf{D}_{\Psi}(\pi_1\|\mu) + \mathsf{D}_{\Psi}(\pi_2\|\nu), \label{formulation:UOT_sup}
%\end{align}

In the UOT problem \cref{formulation:UOT_sup}, the transport cost is finite only when the transport map $\bar{\pi}$ has the marginal distribution $\bar{\pi}_1 = a \mu$ and $\bar{\pi}_2 = a \nu$, which satisfies $\frac{d \bar{\pi}_1}{d \mu} = \frac{d \bar{\pi}_2}{d \nu} = a$. Thus, $\mathsf{D}_{\Psi}(\pi_1\|\mu) + \mathsf{D}_{\Psi}(\pi_2\|\nu) = C$ is a constant.

As a result, finding the optimal unbalanced transport map for \cref{formulation:UOT_sup} is equivalent to find
%$$\argmin_{\bar{\pi} \in \gM(\gX\times \gX)}\int c(x,y) d\bar{\pi}(x,y)$$

%Or:
\begin{align*}
\argmin_{\pi \in \Pi(a \mu, a \nu)}\int c(x,y) d\pi(x,y)
\end{align*}
We also have
\begin{align*}\min_{\pi \in \Pi(a \mu, a \nu)}\int c(x,y) d\pi(x,y) = \OTsf(a \mu, a \nu) = a \OTsf(\mu, \nu).
\end{align*}

% Then: $\UOTsf(\mu,\nu) = \OTsf(a \mu, a \nu) = a \OTsf(\mu, \nu)$.

Let $\bar{\pi}^* = a \pi^*$, we have:
$$\int c(x,y) d\bar{\pi}^*(x,y) = a \int c(x,y) d\pi^*(x,y) = a \OTsf(\mu, \nu)$$

Therefore, $a \pi^*$ is the optimal transport map of the UOT problem (Q.E.D). Lastly, we will \textbf{explain intuitively why using Softplus can filter out abnormal data}.

First, using the Remark in this section, given $\Psi^*(x) = \ln (1 + e^x)$, we have:
\begin{equation}
\label{eq:softplus_psi}
    \Psi = \begin{cases}
         x \ln x + (1 - x) \ln (1 - x), & \text {if } x \in (0,1) \\ 
        \infty, & \text {otherwise}
    \end{cases}
\end{equation}

Compared to the penalized linear function (refer to \cref{eq:linear_psi}), the UOT problem with the convex conjugate function of Softplus does not reduce to a normal OT problem.

Assume that  $\mathsf{D}_{\Psi}(\pi_1\|\mu) + \mathsf{D}_{\Psi}(\pi_2\|\nu)$  attains its minimum at $\zeta$
(\cref{formulation:UOT_sup}), then if  $\frac{d \pi_1}{d \mu} = \frac{d \pi_2}{d \nu} = \zeta$, then it reduces the UOT problem to an OT problem. 

However, if there are outliers, which means that the transportation costs at some locations are very large, then one can decrease mass at those locations of  $\pi$ so that the change of $\mathsf{D}_{\Psi}(\pi_1\|\mu) + \mathsf{D}_{\Psi}(\pi_2\|\nu)$ is much smaller than the decrease in total transportation cost  $\int c(x,y) d\pi(x,y)$. It explains why both KL and Softplus have the ability to filter out outliers. 

\begin{figure}
\centering
\subfloat[KL]{
    \includegraphics[width=0.45\linewidth]{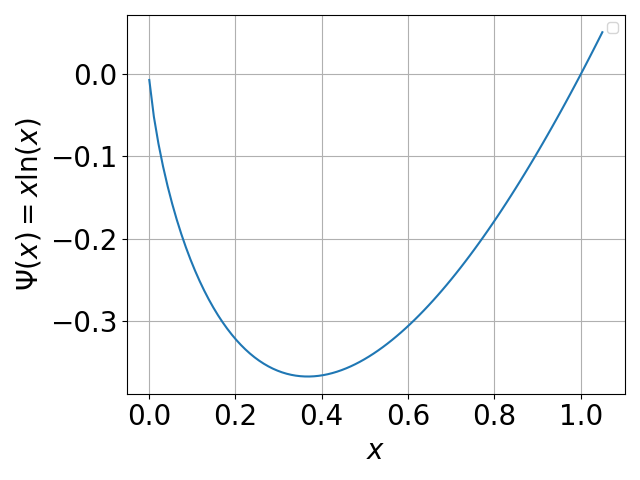}
    % \label{fig:stl10}
}
% ~
\subfloat[Softplus]{
    \includegraphics[width=0.45\linewidth]{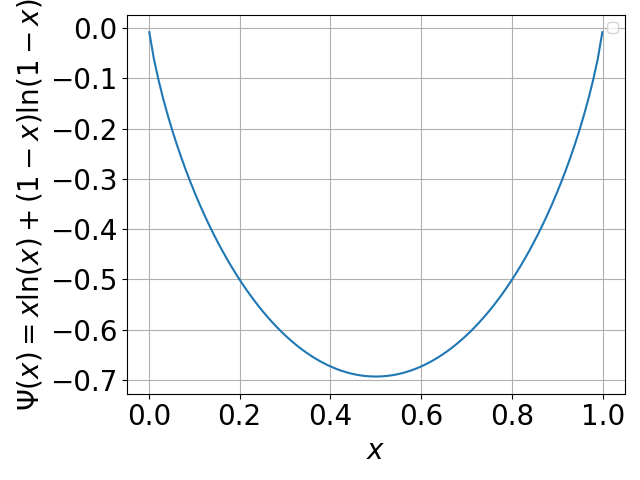}
    % \label{fig:cifar10}
}
\caption{The graph of KL function and function whose convex conjugate is Softplus.}
\label{fig:graph_Psi_KL_Softplus}
\end{figure}

It is noteworthy that, despite sharing many similarities (\cref{fig:graph_Psi_KL_Softplus}), the convex conjugate functions of these two functions are very different, with Softplus owing some benefits due to its Lipschitz continuity property.

% \section{Discussion on training stability of RDUOT}

\section{Additional Results} \label{appen:C}

\subsection{Toy example}
% \vspace{-1cm}
\begin{figure}[!ht]
    \centering
    \includegraphics[width=\linewidth]{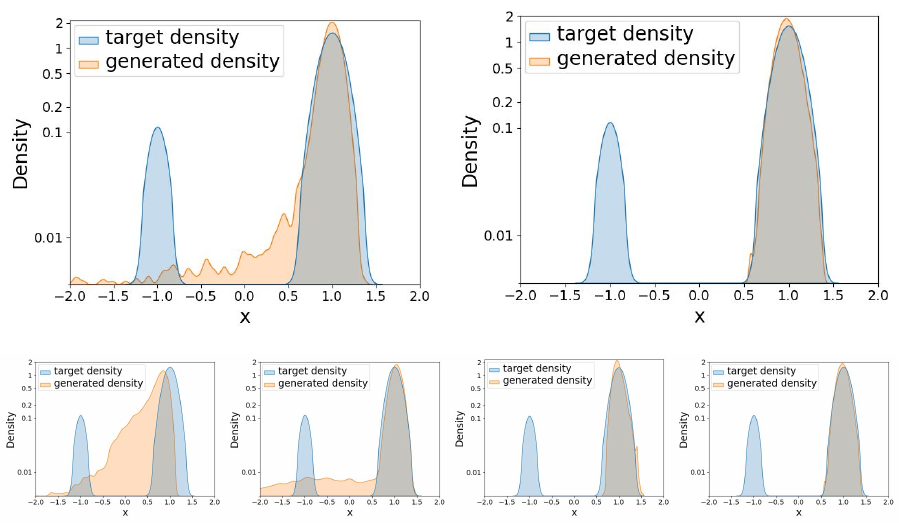}
    \caption{\textbf{Outlier Robustness on Toy Dataset} with $5\%$ outliers. The toy dataset is a mixture of two Gaussians $\mathcal{N}(1, 0.1)$ (clean dataset), $\mathcal{N}(-1, 0.05)$ (outlier dataset) with the mixture rate is ($0.95, 0.05$).  In the first row, subplots compare target and generated densities between DDGAN and RDUOT. Left: DDGAN; Right: RDUOT. The second row showcases partial timestep RDUOT results. From left to right, semi-dual UOT loss is applied to the first 1, 2, 3 timesteps, and then to all timesteps.}
    \label{fig:toy}
% \vspace{-6mm}
\end{figure}

To demonstrate the effectiveness of our RDUOT method on corrupted datasets, we initially compare the generated density obtained by training RDUOT and DDGAN techniques with the ground truth target density on a toy dataset. As illustrated in \cref{fig:toy}, we visually observe that RDUOT exclusively generates new data points that align with the clean mode on the right, whereas DDGAN produces outlier data scattered between the two modes. 

As DDGAN comprises multiple diffusion steps that are trained with adversarial networks, a natural question arises: How well does RDUOT perform when only partially applying the proposed loss within the DDGAN framework? Referring to \cref{fig:toy}, it becomes evident that the performance of RDUOT with partial timesteps falls behind that of RDUOT with all timesteps. We referred to "RDUOT with all timesteps" simply as "RDUOT" in other sections of this paper.

% \subsection{Traditional Unsupervised Learning as Preprocessing mechanism}

% When dealing with a perturbed dataset, the simple traditional approach is to preprocess it to remove the outliers. The approach can be used in parallel with RDUOT for better outlier robustness. As presented in \cref{sec_experiment}, RDUOT can even outperform DDGAN in clean datasets (\cref{tab:FID_clean_cifar},\cref{tab:quality_cifar10}, \cref{tab:stl10}). Therefore, even if we can remove all outliers, using RDUOT instead of DDGAN can provide better performance. For the dataset of CELEBAHQ $64 \times 64$ perturbed by $5\%$ of Fashion-MNIST images, we apply isolation forest \cite{liu2008isolation} for preprocessing, which raises the clean ratio to $97\%$. We then train DDGAN on the preprocessed data and achieve the FID of $11.3$, which is significantly worse than RDUOT trained on raw data with FID of $\textbf{7.89}$.

\subsection{Traditional Unsupervised Learning as Preprocessing Mechanism} \label{appen:preprocess}

\begin{figure}
    \centering
    \includegraphics[width=0.5\linewidth]{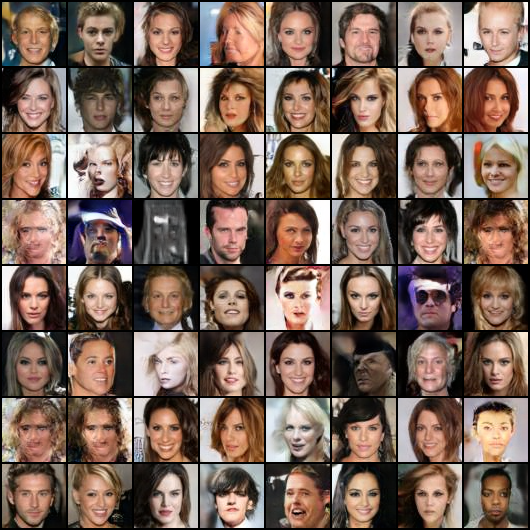}
    \caption{Isolation forest + DDGAN}
    \label{fig:isolation_forest_ddgan}
\end{figure}

When dealing with a perturbed dataset, the simple traditional approach is to preprocess it to remove the outliers. The approach can be used in parallel with RDUOT for better outlier robustness. As presented in \cref{sec_experiment}, RDUOT can even outperform DDGAN in clean datasets (\cref{tab:FID_clean_cifar},\cref{tab:quality_cifar10}, \cref{tab:stl10}). Therefore, even if we can remove all outliers, using RDUOT instead of DDGAN can provide better performance. For the dataset of CELEBAHQ $64 \times 64$ perturbed by $5\%$ of Fashion-MNIST images, we apply isolation forest \footnote{Liu, F.T., Ting, K.M., Zhou, Z.H.: Isolation forest. In: 2008 eighth ieee international conference on data mining. pp. 413–422. IEEE (2008)} for preprocessing, which raises the clean ratio to $97\%$. We then train DDGAN on the preprocessed data and achieve the FID of $11.3$ (qualitative result in \cref{fig:isolation_forest_ddgan}), which is significantly worse than RDUOT trained on raw data with FID of $\textbf{7.89}$.

% When it comes to a perturbed dataset, one can think of preprocessing it by applying machine learning techniques to remove the outliers. This does not make any conflicts with or degrade the motivation and diminish the significance of RDUOT. On the contrary, it can be applied concurrently with RDUOT. As presented in \cref{sec_experiment}, RDUOT can even outperform DDGAN in clean datasets. So even if one can remove all outliers perfectly, it is still worth considering using RDUOT instead of DDGAN. 

% Also, one should be concerned about some of these problems when applying techniques to remove the outliers. It takes resources to find good outlier detection techniques. In terms of high-dimensional data, like images, the traditional clustering models have low performance due to the curse of dimensionality \footnote{Radovanovic, M., Nanopoulos, A., Ivanovic, M.: Hubs in space: Popular nearest neighbors in high-dimensional data. Journal of Machine Learning Research 11(sept), 2487–2531 (2010)}.%\cite{radovanovic2010hubs}.
% Thus, we tried some popular anomaly detection techniques on the dataset of CelebHQ 64 × 64 perturbed by 5\% of Fashion-MNIST images and achieved the highest accuracy of 97,31\% using isolation forest \footnote{Liu, F.T., Ting, K.M., Zhou, Z.H.: Isolation forest. In: 2008 eighth ieee international conference on data mining. pp. 413–422. IEEE (2008)}.%\cite{liu2008isolation}.
% DDGAN has an FID score of 11.32 with this filtered dataset, which is still worse than pure RDUOT. \cref{fig:isolation_forest_ddgan} shows the results of this approach. 

\subsection{Ablation on number of time steps and cost weight $\tau$}
In this section, we first discuss the performance of RDUOT with different timesteps. As shown in \cref{table:num_timesteps}, we observe that our proposed method outperforms DDGAN with different numbers of timesteps. It is worth noting that our method still works well as the number of timesteps increases to $8$. In contrast, DDGAN with $8$ timesteps witnesses a performance decrease.
\begin{table}[]
    \centering
  \footnotesize
    \begin{tabular}{l|ccc}
         \toprule
          $\#$ of timesteps &  2 &  4 & 8 \\
         \midrule
          RDUOT  & 3.84  & 2.95  & \underline{\textbf{2.65}}   \\
          DDGAN  & 4.08  & \underline{3.75} & 4.36 \\
         \bottomrule
    \end{tabular}
    \caption{RDUOT and DDGAN with different numbers of training timesteps on clean CIFAR10 dataset}
    \label{table:num_timesteps}
\end{table}

As can be seen in \cref{alg:SUOT}, an RDUOT model training on a four-timestep setting can still generate images ($\hat{x}_0$) with fewer than four timesteps. We measured the performance of the proposed model in each scenario on the clean CIFAR10 dataset and reported the result in \cref{table:fewer_timesteps_generation}. We observe that the more generation timesteps, the higher the quality of the generated samples.

\begin{table}[]
    \centering
  \footnotesize
    \begin{tabular}{l|cccc}
         \toprule
          \# of sampling NFE &  1 &  2 & 3 & 4 (full) \\
         \midrule
          FID  & 49.53  &  13.32 & 3.44 & 2.95 \\
         \bottomrule
    \end{tabular}
    \caption{FID of RDUOT sampling with fewer NFEs (skip the later steps) on clean CIFAR10 dataset.}
    \label{table:fewer_timesteps_generation}
\end{table}

Then, we perform the experiment with different $\tau$ values. The $\tau$ value is an important hyperparameter that is in charge of filtering outliers. From \cref{table:cost_weight_tau}, we see that when $\tau$ is too small ($\leq$1e-3), our model is unable to filter the outlier leading to low precision compared to clean data. However, when $\tau$ is too high, our model could wrongly filter the data leading to low FID.

\begin{table}[]
    \centering
  \footnotesize
    \begin{tabular}{l|ccccc}
         \toprule
         $\tau$ &  1e-4 &  3e-4 & 1e-3 & 2e-3 & 5e-3 \\
         \midrule
          FID  & 6.74  &  6.09 & 4.37 & \textbf{3.94} & 5.98 \\
         \bottomrule
    \end{tabular}
    \caption{FID of RDUOT with different $\tau$ on CI+MT dataset}
    \label{table:cost_weight_tau}
\end{table}

\subsection{Other Qualitative Results}
\begin{figure}
\centering
\subfloat[DDGAN STL-10]{
    \includegraphics[width=0.45\linewidth]{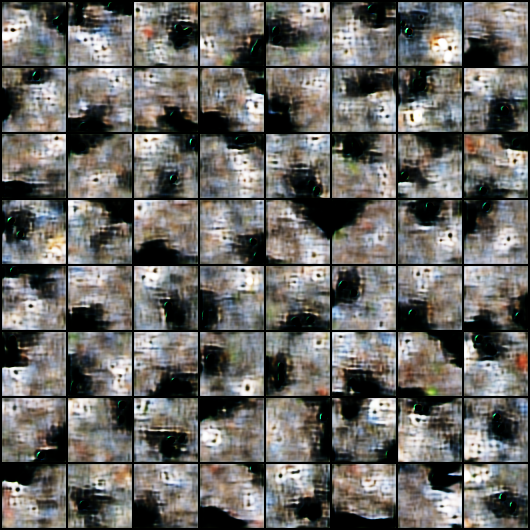}
    % \label{fig:stl10}
}
% ~
\subfloat[RDUOT STL-10]{
    \includegraphics[width=0.45\linewidth]{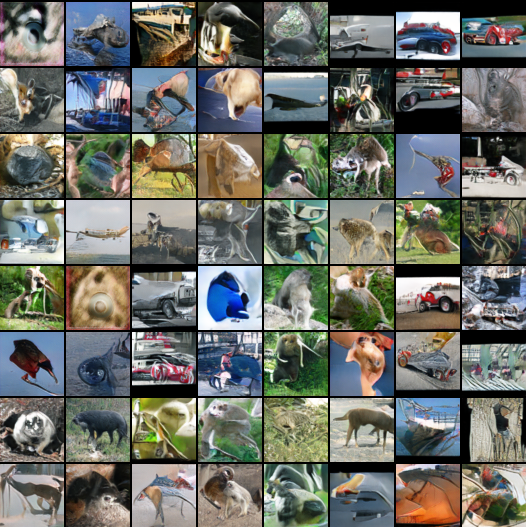}
    % \label{fig:cifar10}
}
\caption{Qualitative comparison of RDUOT and DDGAN on STL-10 at epoch 300. RDUOT converges faster than DDGAN.}
\label{fig:stl10_300_step}
\end{figure}

\begin{figure}
    \centering
    \includegraphics[width=0.8\linewidth]{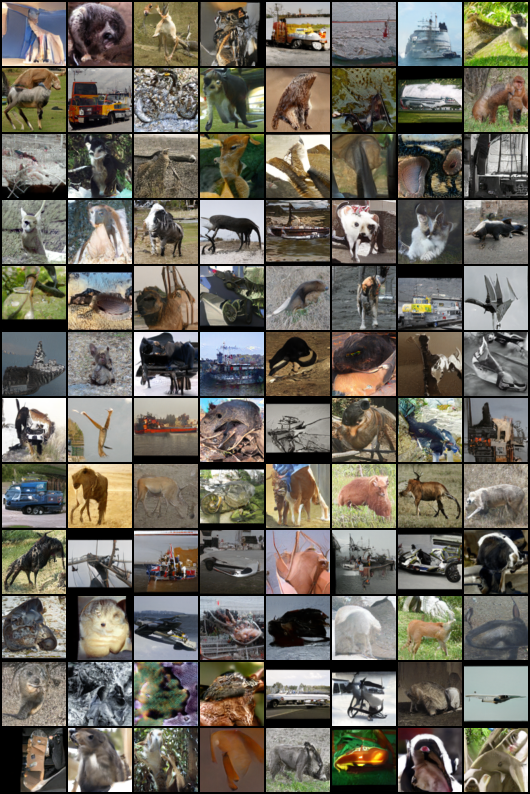}
    \caption{Non-curated STL-10 qualitative images.}
    \label{fig:non_curated_stl10}
\end{figure}

\begin{figure}
    \centering
    \includegraphics[width=0.8\linewidth]{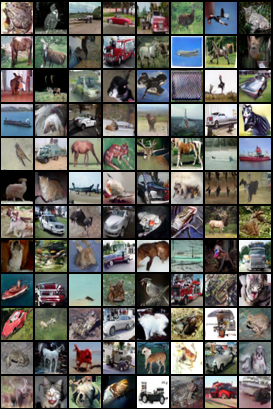}
    \caption{Non-curated CIFAR-10 qualitative images.}
    \label{fig:non_curated_cifar10}
\end{figure}

\begin{figure}
    \centering
    \includegraphics[width=0.8\linewidth]{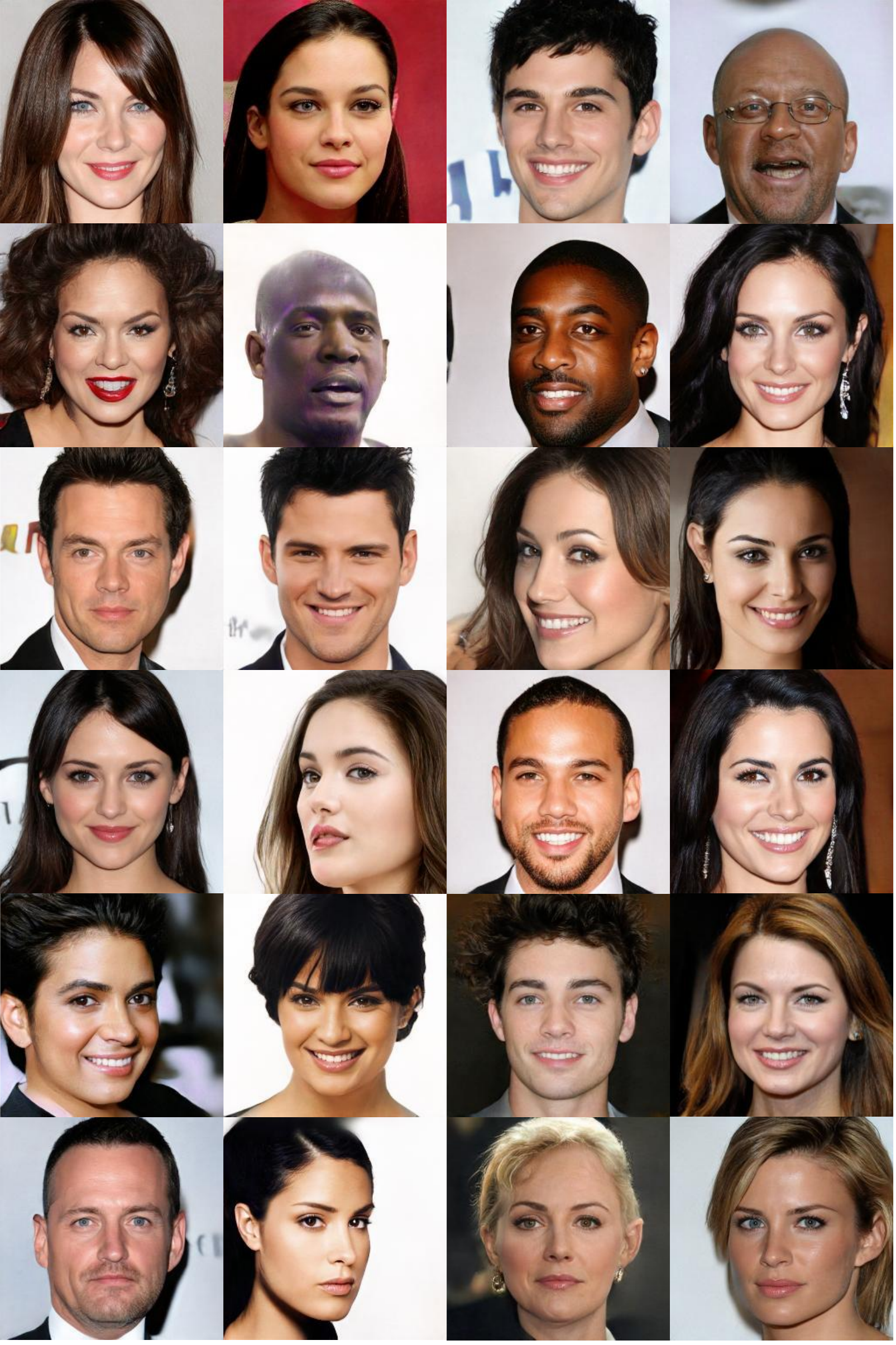}
    \caption{Non-curated CELEBAHQ-256 qualitative images.}
    \label{fig:non_curated_celeb256}
\end{figure}

As discussed in \cref{sec_experiment}, RDUOT converges faster than DDGAN. As can be seen clearly in \cref{fig:stl10_300_step},  at epoch 300, the generated images of RDUOT have much higher qualities compared to those of DDGAN. We show the non-curated qualitative figure of RDUOT on clean datasets in \cref{fig:non_curated_stl10}, \cref{fig:non_curated_cifar10}, \cref{fig:non_curated_celeb256}.

\begin{figure}
\centering
\subfloat[DDGAN CE+FCE]{
    \includegraphics[width=0.45\linewidth]{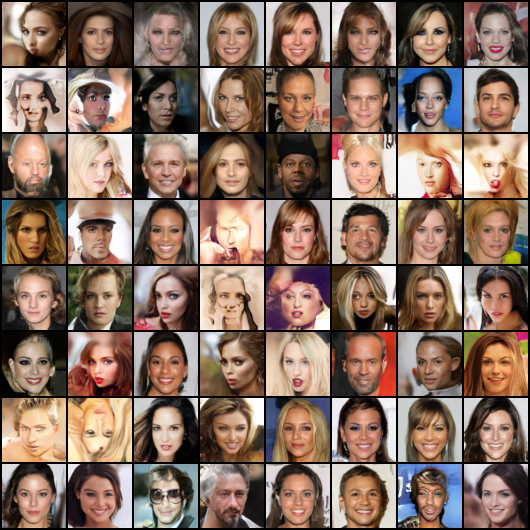}
    % \label{fig:stl10}
}
% ~
\subfloat[RDUOT CE+FCE]{
    \includegraphics[width=0.45\linewidth]{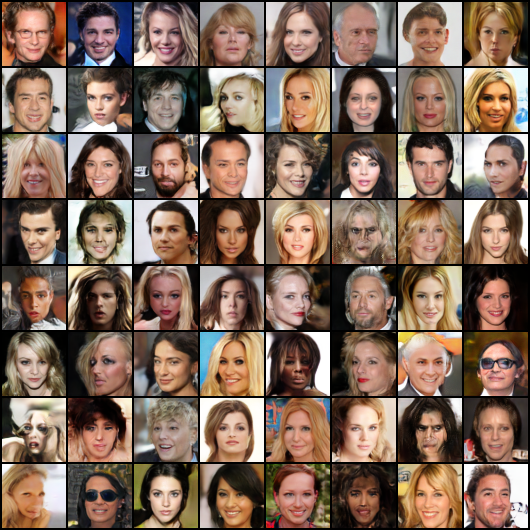}
    % \label{fig:cifar10}
}
\caption{Qualitative comparison of RDUOT and DDGAN on CELEBAHQ $64 \times 64$ perturbed by vertical flip outliers.}
    \label{fig:qualitative_fce}
\end{figure}

\cref{fig:qualitative_fce} represents the results of DDGAN and RDUOT on CELEBAHQ $64 \times 64$ perturbed by VERTICAL FLIP  CELEBAHQ $64 \times 64$. Though the difference between FID scores is just $0.5$ as in \cref{table:ortherdataset}, the generated images of DDGAN are much more vulnerable to flip face property.

\end{document}